%% file: main.tex
\documentclass[journal]{IEEEtran}
\IEEEoverridecommandlockouts
\usepackage{graphicx}
\def\BibTeX{{\rm B\kern-.05em{\sc i\kern-.025em b}\kern-.08em
    T\kern-.1667em\lower.7ex\hbox{E}\kern-.125emX}}

\usepackage{blindtext}
\usepackage{caption}

\let\oldtwocolumn\twocolumn
\renewcommand\twocolumn[1][]{%
    \oldtwocolumn[{#1}{
    \begin{center}
    \vskip-7ex
        \centering
        \includegraphics[
        width=0.98\textwidth,
        trim={10 0 10 10mm}, 
        clip
    ]{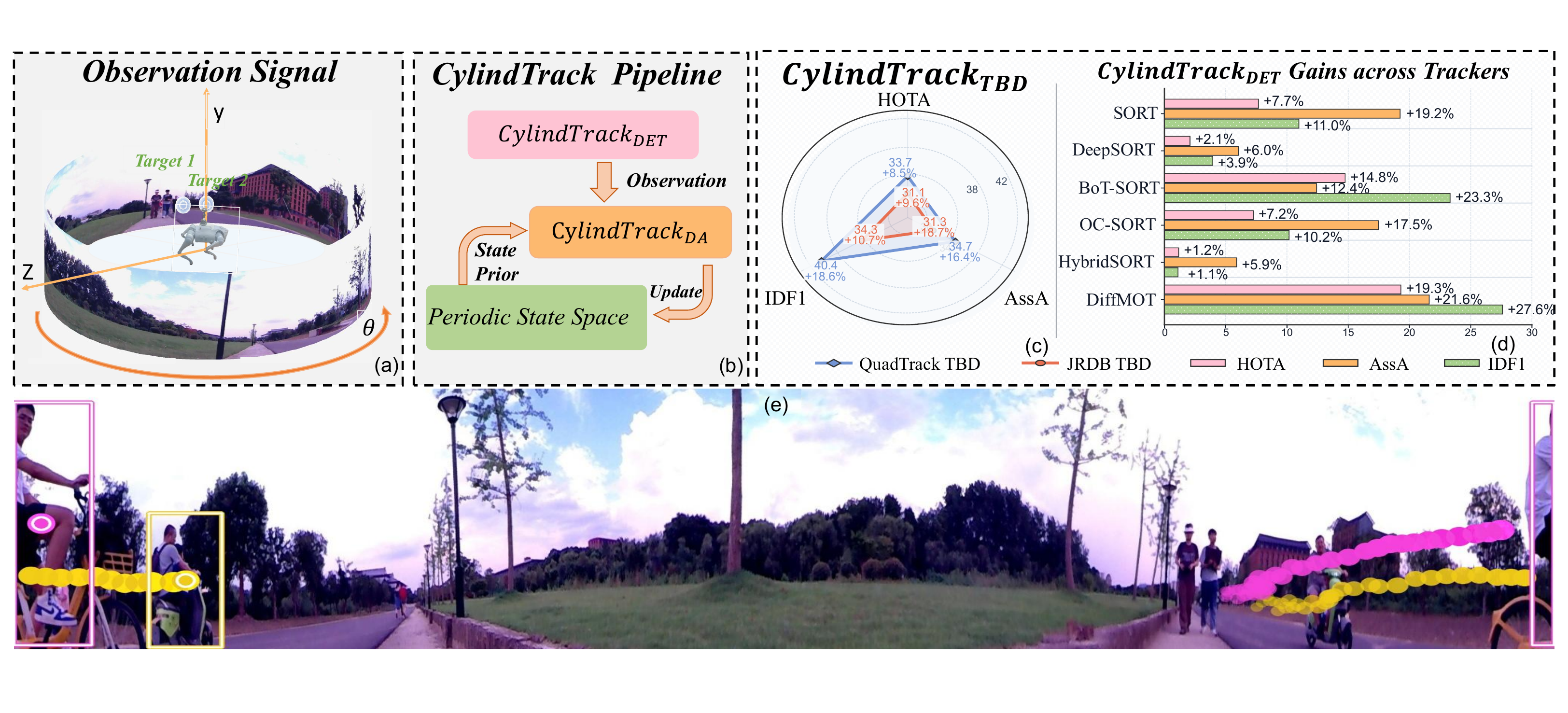}
        \vskip-5ex
        \captionof{figure}{Overview of our proposed CylindTrack framework for panoramic multi-object tracking.
        (a)–(b) CylindTrack extends conventional planar tracking by incorporating cylindrical topology and depth-aware temporal trajectory optimization into the standard tracking-by-detection paradigm.
        It performs motion prediction and trajectory association in a depth-aware panoramic cylindrical space, while preserving the modular structure of online TBD trackers.
        (c) CylindTrack outperforms representative TBD baselines on JRDB~\cite{martin2021jrdb} and QuadTrack~\cite{luo2025omnitrack} benchmarks, demonstrating the effectiveness of depth-aware panoramic motion modeling and spatio-temporal consistency.
        (d) The proposed depth-enhanced detector with geometric-temporal consistency yields general improvements of average identity-oriented metrics~\cite{bewley2016sort,wojke2017deepsort,yang2024hybrid,lv2024diffmot,aharon2022botsortrobustassociationsmultipedestrian} on JRDB and QuadTrack.
        (e) When targets move across panoramic image boundaries, CylindTrack preserves angular trajectory continuity and explicitly handles cross-boundary association, improving identity preservation and reducing trajectory fragmentation.
        }
        \label{fig:teaser}
    \end{center}
    }]
}
\vspace{-4ex}

\usepackage{amsmath}
\usepackage{amssymb}

\usepackage[utf8]{inputenc}
\usepackage[T1]{fontenc}

\usepackage{booktabs}
\usepackage{array}
\usepackage{float}
\usepackage{multirow}
\usepackage{siunitx}

\usepackage[table]{xcolor}
\definecolor{rblue}{rgb}{0,0.5,1}
\definecolor{awesome}{rgb}{1.0, 0.13, 0.32}
\definecolor{hollywoodcerise}{rgb}{0.96, 0.0, 0.63}
\definecolor{lasallegreen}{rgb}{0.03, 0.47, 0.19}
\definecolor{hanpurple}{rgb}{0.32, 0.09, 0.98}

\usepackage{pifont}

\usepackage[ruled,vlined,linesnumbered]{algorithm2e}
\usepackage{cite}
\usepackage{setspace}
\newcommand{\cmark}{\ding{51}}
\newcommand{\xmark}{\phantom{\ding{51}}}

\newcommand{\best}[1]{\textbf{#1}}

\newcommand{\eid}[1]{%
  \raisebox{-0.05ex}{\scalebox{1.22}{\ding{\numexpr171+#1\relax}}}%
}

\usepackage[
    pagebackref=false,
    breaklinks=true,
    colorlinks,
    bookmarks=false
]{hyperref}

\hypersetup{
    colorlinks=true,
    linkcolor={red},
    citecolor={hanpurple},
    urlcolor={magenta}
}
\usepackage{icomma}

\usepackage{marvosym}

\begin{document}
\title{CylindTrack: Depth-Aware Cylindrical Motion Modeling for Panoramic Multi-Object Tracking in Mobile Robots}

\author{Buyin Deng$^{1,*}$, Kai Luo$^{1,*}$, Lingxin Huang$^{1}$, Xinqi Liu$^{1}$, Fei Cheng$^{2,3}$, Hang Zheng$^{4}$,\\Liming Yin$^{3}$, and Kailun Yang$^{1,\dag}$%
\thanks{This work was supported in part by the National Natural Science Foundation of China (Grant No. 62388101 and No. 62473139), in part by the Hunan Provincial Research and Development Project (Grant No. 2025QK3019), and in part by the State Key Laboratory of Autonomous Intelligent Unmanned Systems (the opening project number ZZKF2025-2-10).}%
\thanks{$^{1}$The authors are with the School of Artificial Intelligence and Robotics and the National Engineering Research Center of Robot Visual Perception and Control Technology, Hunan University, China (email: kailun.yang@hnu.edu.cn).}%
\thanks{$^{2}$The author is with the School of Advanced Technology, Xi'an Jiaotong-Liverpool University, China (fei.cheng@xjtlu.edu.cn).}%
\thanks{$^{3}$The authors are with Suzhou VSDeep Intelligent Technology Co., Ltd., China (email: lm.yin@vsdeep.com).}%
\thanks{$^{4}$The author is with BWTON Technology Co., Ltd., China (email: zhenghang@bwton.com).}%
\thanks{$^{*}$Equal contribution.}%
\thanks{$^{\dag}$Corresponding author: Kailun Yang.}
}

\maketitle

\input{body/abstract}
\begingroup 
\renewcommand{\abstractname}{Note to Practitioners} 
\begin{abstract} 
This work targets mobile robots that need to maintain consistent identities of surrounding targets in panoramic video. When a target crosses the left-right seam of an equirectangular image, conventional planar tracking can interrupt the trajectory or assign a new identity. CylindTrack addresses this issue by combining periodic motion modeling with temporally filtered relative-depth cues, helping maintain consistent observation histories across the panorama seam. Its tracking-by-detection architecture separates observation generation from temporal association, facilitating integration into modular robot perception pipelines. Evaluation on representative robot-captured panoramic benchmarks demonstrates improvements in identity consistency under the tested conditions. Deployment on specific platforms requires assessing end-to-end latency, computational resource requirements, and tracking stability under sustained robot motion. Relative depth serves as an association cue rather than a calibrated distance measurement; applications requiring metric geometry, such as collision avoidance, therefore need complementary sensing or estimation. Under severe occlusion or seam-fragmented detections, maintaining trajectory continuity also depends on the availability and consistency of object observations. Integrating seam-consistent detection and camera-motion-aware state estimation offers a direction for further improving persistent panoramic tracking in dynamic robotic environments.
\end{abstract} 
\endgroup 

\IEEEpeerreviewmaketitle

\input{body/introduction}
\input{body/related_works}
\input{body/method}
\input{body/experiment}
\input{body/conclusion}
\bibliographystyle{IEEEtran}
\bibliography{reference}

\vfill

\end{document}

%% file: body/abstract.tex
\begin{abstract}
Maintaining consistent target identities and observation histories is essential for mobile robot perception. Panoramic Multi-Object Tracking (MOT) benefits from broad visual coverage but must address coordinate discontinuities at the equirectangular image seam and fluctuations in relative-depth estimates. Such fluctuations can undermine depth-aware association, highlighting the need for temporally consistent depth cues. We propose CylindTrack, a depth-aware cylindrical tracking-by-detection framework that jointly models panoramic topology and temporal depth consistency. Depth-Temporal Trajectory Modeling (DTM) maintains instance depth and its rate of change as filtered trajectory states. Spherical Spatio-Temporal Consistency Learning (SSTC) combines a Temporal Mixer with Spherical Geometry-aware Attention to refine depth-aware representations. A Topology-Aware Cylindrical Motion Model (TCMM) represents horizontal motion in an unwrapped angular state and incorporates periodic overlap, angular consistency, and filtered depth cues into association. Experiments on QuadTrack and JRDB demonstrate improved identity consistency and trajectory continuity, with HOTA scores of 33.67 and 31.12 and IDF1 scores of 40.45 and 34.33, respectively. The framework processes 28.56 and 21.34 frames per second, respectively, on an NVIDIA GeForce RTX 3090 GPU under the reported evaluation settings. These results support persistent panoramic multi-object tracking for mobile robot perception. The source code will be released at \url{https://github.com/warriordby/CylindTrack}.
\end{abstract}

%% file: body/introduction.tex
\section{Introduction}

Mobile robots operating in dynamic environments require continuous perception of surrounding objects to support autonomous navigation, interaction, and scene understanding~\cite{martin2021jrdb,shenoi2020jrmotrealtime3dmultiobject}. 
Multi-Object Tracking (MOT) serves as a perception component in these automated systems by maintaining object identities and linking successive observations into coherent trajectories~\cite{wang2024sttracknet,yan2026chtracker,muzaddid2024ntrack}.
This temporal representation supports persistent monitoring of multiple object categories and provides consistent motion histories for downstream decision-making. Panoramic cameras offer broad spatial coverage for this task, but reliable tracking depends on both the quality of individual observations and the consistency of motion prediction and data association across frames~\cite{deng2025deptrmot,liu2025sparsetrack,cui2026depthsort}. This motivates tracking representations that account for panoramic imaging geometry while maintaining temporally consistent association cues.

Panoramic cameras offer a $360^{\circ}$ horizontal field of view (FoV), reducing directional blind spots and providing broad spatial context for surrounding perception~\cite{gao2022review,shi2023panoflow}. Exploiting this coverage requires processing methods that account for panoramic observation characteristics. For example, distortion-aware convolution incorporates the image-forming principle into panoramic semantic segmentation~\cite{9667494}, while PanoGlassNet addresses wide-FoV glass detection using panoramic RGB and intensity images and deformable contextual features~\cite{10504145}. These studies motivate the use of observation-specific representations in panoramic visual processing. For tracking, the corresponding challenge extends beyond per-frame feature extraction to maintaining consistent target states and identities across successive observations~\cite{liu2018simple_online,he2021know_surroundings,luo2025omnitrack}.

The Tracking-By-Detection (TBD) paradigm offers a modular framework for this task by separating frame-wise detection from temporal association~\cite{bewley2016sort,wojke2017deepsort,zhang2022bytetrack,cao2023observation}. Detections provide object observations, and the tracker combines them with predicted trajectory states to maintain identities over time. This separation permits observation refinement and association modeling to be developed within a common pipeline. Related MOT studies have explored spatio-temporal detection representations~\cite{wang2024sttracknet} and confidence-guided hierarchical association~\cite{yan2026chtracker} to improve data association. However, conventional image-plane motion models and overlap measures do not explicitly account for the periodic horizontal domain of equirectangular projection (ERP). Moreover, depth-assisted association depends on the temporal consistency of the geometric cues provided by the detector~\cite{deng2025deptrmot}. These issues create two coupled challenges for panoramic TBD.

First, the tracking state and association measures must account for the topology of the observation domain. The left and right edges of an ERP image represent adjacent viewing directions, although they are separated by almost the full image width in planar coordinates. 
As illustrated in Fig.~\ref{fig:paradigm}, a target crossing the panorama seam can exhibit an abrupt horizontal-coordinate jump despite continuous physical motion. Directly applying planar displacement estimation and bounding-box overlap can therefore corrupt velocity prediction and weaken association near the $0^{\circ}/360^{\circ}$ seam~\cite{liu2018simple_online,luo2025omnitrack,liu2026s3kf}. The resulting discontinuity originates in the coordinate representation and should be addressed consistently in both motion prediction and association.

\input{figures/paradigm}
Second, geometric cues must remain sufficiently consistent over time to support identity discrimination. Occlusion, close target interactions, and seam-separated object observations can make bounding boxes and appearance cues ambiguous. Moreover, incomplete annotations in these challenging regions, as illustrated in Fig.~\ref{fig:pano_challenge}, limit the ability of tracking models to learn and evaluate continuous identity association under complex panoramic observations. Relative depth can complement two-dimensional localization and appearance cues by providing additional geometric information~\cite{wu2024depthmot,wang2025pd,deng2025deptrmot}. However, depth estimation in panoramic videos poses additional challenges, as camera motion and nonlinear image motion complicate temporal modeling~\cite{song2026pvdepth}. Fluctuations in frame-wise monocular depth estimates can make direct association based on depth differences sensitive to observation noise and scale variation~\cite{video_depth_anything}. Moreover, the sensitivity of depth predictions to spherical spatial transformations~\cite{cao2025panda} can further undermine the temporal consistency of the depth cues used for association. Therefore, reliable depth-assisted association requires temporal modeling of depth cues along each trajectory.

To address these challenges, we propose \textbf{CylindTrack}, a depth-aware cylindrical TBD framework that couples observation refinement with topology-consistent trajectory estimation. We first introduce \textbf{Depth-Temporal Trajectory Modeling (DTM)}, which maintains instance depth and its temporal rate as a compact, filtered trajectory state. Depth is used here as a relative geometric cue for association, without assuming calibrated physical range. We then develop \textbf{Spherical Spatio-Temporal Consistency Learning (SSTC)}, which combines a query-based Temporal Mixer with Spherical Geometry-aware Attention to encourage temporal coherence and geometric alignment in the detector's depth-aware representations. Finally, a \textbf{Topology-Aware Cylindrical Motion Model (TCMM)} maintains horizontal motion in an unwrapped angular state and combines horizontal-periodic overlap, angular consistency, and filtered depth cues to enhance trajectory association. The framework thereby addresses observation consistency and trajectory-state continuity within a unified panoramic tracking pipeline, supporting persistent multi-object monitoring for mobile robot perception.

\input{figures/pano_challenge}

We evaluate CylindTrack on the QuadTrack~\cite{luo2025omnitrack} and JRDB~\cite{martin2021jrdb} panoramic MOT benchmarks. Across seven representative TBD trackers, replacing the DepTR-MOT detector~\cite{deng2025deptrmot} with the SSTC-enhanced \(\mathrm{CylindTrack}_{\mathrm{Det}}\) increases the mean IDF1 score by $11.83\%$ on QuadTrack and $6.15\%$ on JRDB; the corresponding relative increases in mean AssA are $8.22\%$ and $11.59\%$, computed from the mean scores across the seven trackers. The complete framework achieves HOTA scores of $33.674$ and $31.117$, AssA scores of $34.665$ and $31.347$, and IDF1 scores of $40.446$ and $34.331$ on QuadTrack and JRDB, respectively, at processing rates of $28.56$ and $21.34$ frames per second. We also introduce Boundary Crossing Identity Consistency (BCIC) to quantify identity preservation during panorama-seam transitions and characterize local tracking robustness beyond conventional sequence-level metrics.

Our main contributions are summarized as follows:
\begin{itemize}
\item We propose \textbf{CylindTrack}, a depth-aware cylindrical panoramic MOT framework that preserves the modular tracking-by-detection pipeline while explicitly modeling the cyclic topology and depth dynamics of equirectangular panoramic videos.

\item We enhance the trajectory association cues with \textbf{Depth-Temporal Trajectory Modeling (DTM)} and \textbf{Spherical Spatio-Temporal Consistency (SSTC)}, which model instance depth as a temporally filtered trajectory-level state and enforce temporal coherence via panoramic geometric alignment. These designs provide more reliable depth-aware cues for trajectory association under occlusion, close interactions, and boundary-crossing motions.

\item We introduce a \textbf{Topology-Aware Cylindrical Motion Model (TCMM)} that represents horizontal target motion in a continuous angular state space and performs topology-consistent association using horizontal-periodic overlap and angular consistency. This mitigates discontinuous motion prediction and unreliable IoU matching near the $0^{\circ}/360^{\circ}$ panorama seam.

\item Extensive experiments on QuadTrack~\cite{luo2025omnitrack} and JRDB~\cite{martin2021jrdb} show that \(\mathrm{CylindTrack}_{\mathrm{Det}}\) improves identity-oriented metrics on average across seven representative trackers over the DepTR detector baseline, while the complete \(\mathrm{CylindTrack}_{\mathrm{TBD}}\) framework outperforms OmniTrack and DepTR-MOT, achieving new state-of-the-art performance on both benchmarks and supporting more reliable temporal association of panoramic sensor observations for target monitoring.
\end{itemize}

%% file: figures/paradigm.tex
\begin{figure}[t]
    \centering
    \includegraphics[
        width=0.48\textwidth,
        trim=230 80 230 20,
        clip
    ]{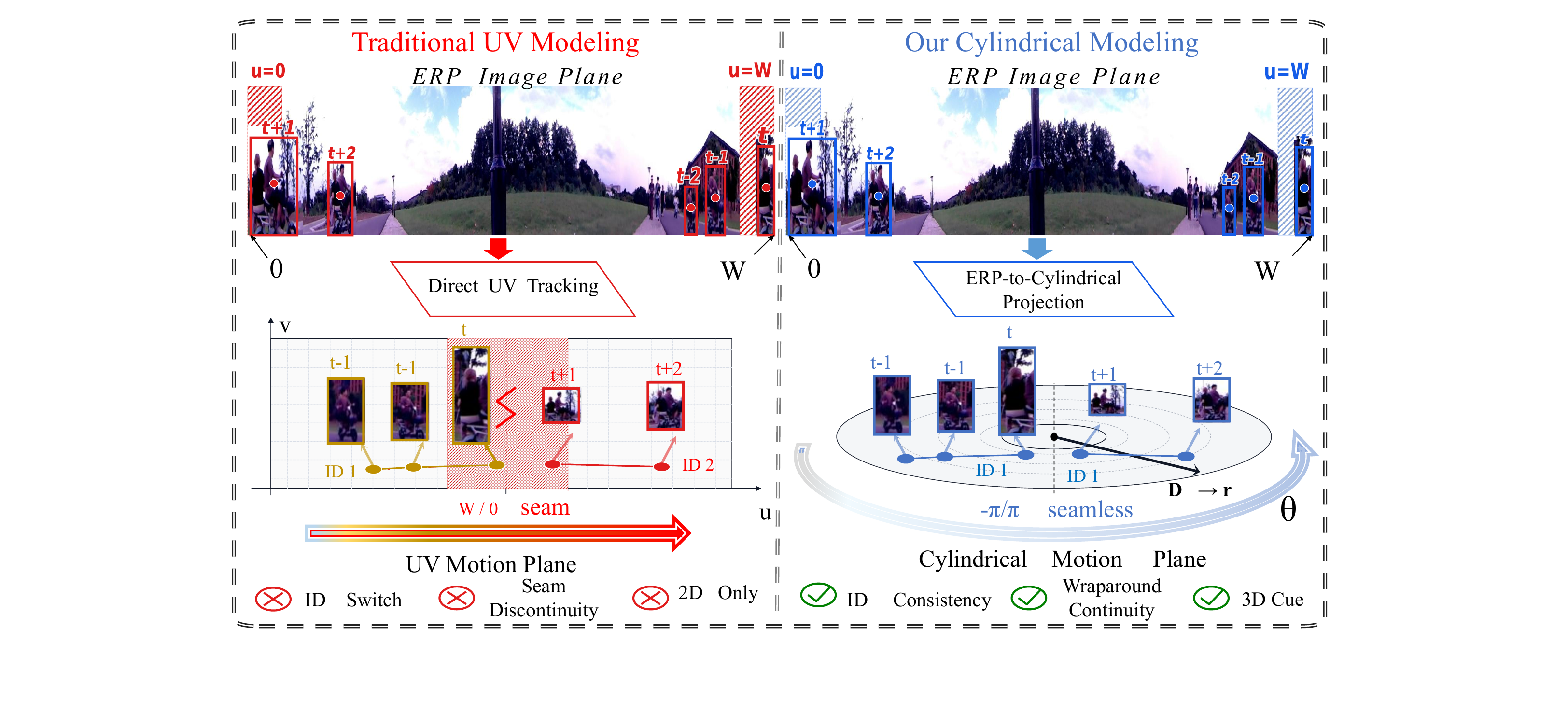}
    \vspace{-1ex}
    \caption{Visualization of boundary-crossing challenges in panoramic multi-object tracking from $t{-}2$ to $t{+}2$, where $t$ denotes the boundary-crossing moment. 
    Traditional UV modeling breaks trajectories at the ERP seam~\cite{luo2025omnitrack,zhang2022bytetrack}, whereas our cylindrical modeling preserves trajectory continuity through topology-consistent motion states, enabling consistent cross-boundary association under periodic spatial transformations.}
    \label{fig:paradigm}
\end{figure}

%% file: figures/pano_challenge.tex
\begin{figure}[t]
    \centering
    \includegraphics[
        width=0.48\textwidth,
    ]{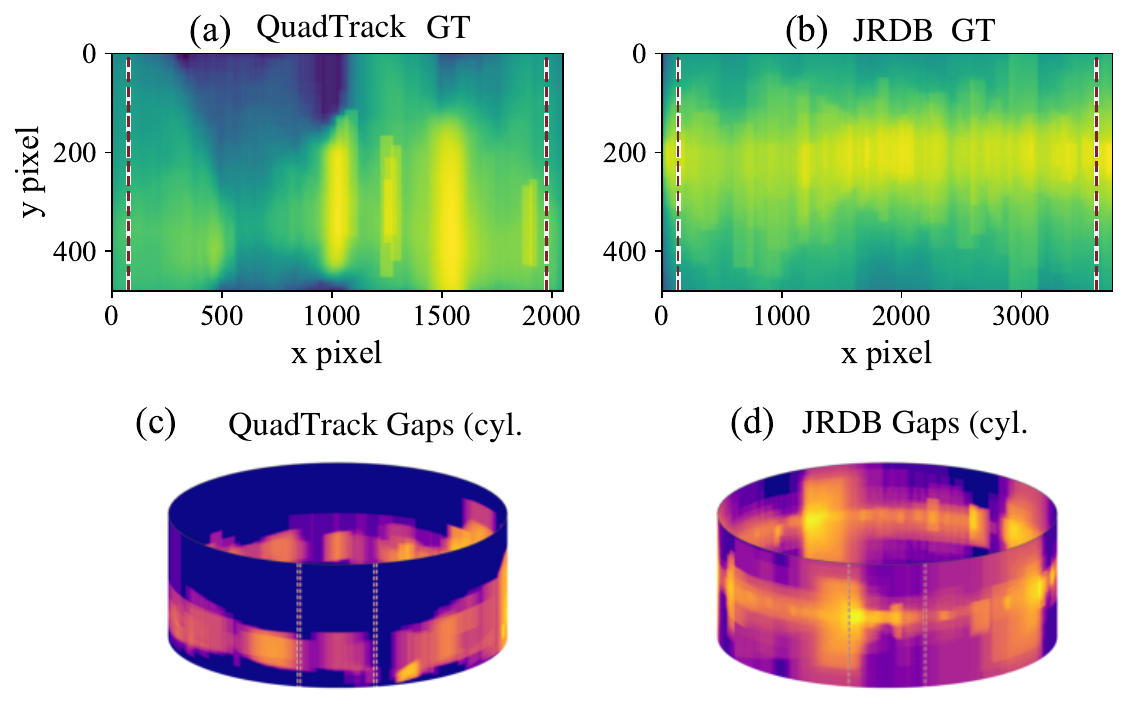}
    \vspace{-1ex}
    \caption{Visualization of boundary annotation distributions and gaps.
(a) and (b) show annotated bounding-box distributions on QuadTrack~\cite{luo2025omnitrack} and JRDB~\cite{martin2021jrdb}, respectively; (c) and (d) show the corresponding missing trajectory annotations.
These gaps mainly occur near the left/right seam and in heavily occluded regions, highlighting annotation sparsity induced by panoramic boundary topology and occlusions.
            }
    \label{fig:pano_challenge}
\end{figure}

%% file: body/related_works.tex
\section{Related Work}

\subsection{Panoramic Scene Understanding}

Panoramic visual perception relies on 360{\textdegree} panoramic images captured by omnidirectional imaging systems, providing continuous observations of the surrounding environment with periodic boundary continuity~\cite{gao2022review,ai2025survey_representation,lin2025one_flight}.
Compared with perspective images, panoramic images exhibit unique geometric distortions, non-uniform spatial sampling, and boundary discontinuities in common projections, motivating task-specific omnidirectional modeling rather than directly applying pinhole-camera pipelines.
Recent studies have advanced panoramic visual understanding across multiple tasks, including panoramic semantic segmentation, \textit{e.g.}, OmniSAM~\cite{Zhong_2025_ICCV} and DAPASS~\cite{chang2026dapss}, panoramic panoptic segmentation~\cite{jaus2023panoramic_panoptic,mei2022waymo,fu2025panopticnerf_360}, and panoramic depth estimation, \textit{e.g.}, PanDA~\cite{cao2025panda} and DA360~\cite{jiang2025da360}. 
These works indicate a broader transition from projection-level image analysis toward holistic omnidirectional scene understanding.

Although panoramic vision naturally benefits Multi-Object Tracking (MOT) through its continuous $360^\circ$ observations, mainstream MOT methods~\cite{bewley2016sort,wojke2017deepsort,zhang2022bytetrack,cao2023observation,di2025hybridtrack,lv2024diffmot} are mainly designed for pinhole-camera inputs and struggle with intrinsic geometric distortions and periodic boundary conditions in omnidirectional views~\cite{liu2018simple_online,he2021know_surroundings,yang2020using_panoramic_videos,fischer2023cc_3dt,yang2026robust}. While OmniTrack~\cite{luo2025omnitrack} has recently introduced a dedicated panoramic MOT framework and the QuadTrack dataset, panoramic MOT in complex unconstrained surroundings still requires more effective motion modeling and trajectory association under the intrinsic geometry and periodic boundary conditions of panoramic images.

\subsection{Kalman-Filter-Based Multi-Object Tracking}
Mainstream two-stage Multi-Object Tracking (MOT) methods typically use a Kalman filter to predict the current state of each trajectory~\cite{bewley2016sort,wojke2017deepsort,cao2023observation}. 
A cost matrix is then constructed based on the distance between the predicted boxes and the detections in the current frame, followed by global assignment using the Hungarian algorithm~\cite{kuhn1955hungarian}. 
Such methods are highly interpretable and modular, as they explicitly incorporate historical motion states while remaining robust to moderate detection noise, making them valuable for practical deployment and analysis~\cite{bewley2016sort,wojke2017deepsort,cao2023observation}. 
Conventional Kalman-filter-based MOT methods usually model target motion in Cartesian image coordinates. 

However, this planar kinematic assumption is inconsistent with the intrinsic topology of panoramic omnidirectional views, leading to boundary-crossing failures and degraded spatio-temporal consistency when targets move across the periodic image boundary~\cite{shen2024multi_object_tracking,liu2018simple_online,yang2020using_panoramic_videos,lo2022depth_aware_spherical,liu2026s3kf}. 
We argue that when the left and right image boundaries are physically connected, the horizontal position should not be modeled as a Euclidean linear coordinate. 
Instead, it should be represented using a non-Euclidean geometric formulation that is consistent with the underlying camera model and panoramic projection geometry~\cite{yang2020using_panoramic_videos,lo2022depth_aware_spherical,liu2026s3kf}.

\subsection{Depth-Informed Multi-Object Tracking}
With the advancement of monocular depth estimation, recent studies have explored depth cues as additional spatial geometric constraints for Multi-Object Tracking (MOT)~\cite{liu2025sparsetrack,wang2025pd,cui2026depthsort,zhao2025detrack,khanchi2025depth_scoring}. 
By capturing relative spatial relationships among targets, depth information is particularly useful for identity disambiguation under occlusion, dense crowds, and close target interactions~\cite{liu2025sparsetrack,wang2025pd,yang2025depth_crowded,peng2025multi_densely_occluded}. 
PD-SORT~\cite{wang2025pd}, SparseTrack~\cite{liu2025sparsetrack}, and DepthSort~\cite{cui2026depthsort} enhance tracking-by-detection pipelines by incorporating pseudo-depth or estimated depth into occlusion reasoning, detection reliability assessment, and data association. 
CAMOT~\cite{limanta2024camot} estimates camera angle and object depth under a common-plane assumption, enabling pseudo-3D association under viewpoint changes. 
DP-MOT~\cite{quach2024depth} introduces subject-ordered depth estimation and an active pseudo-3D Kalman filter to improve association for occluded targets. 
DepthMOT~\cite{wu2024depthmot} jointly performs object detection and scene depth estimation while compensating for irregular camera motion through camera pose estimation. 
ViewTrack~\cite{sun2025view} exploits relative depth relationships inferred from bounding-box geometry to design view-adaptive association strategies. 
DepTR-MOT~\cite{deng2025deptrmot} further proposes a DETR-based detector enhanced with instance-level depth cues, enabling depth-aware target representation and improving association robustness under severe occlusion and close-range interactions.
Beyond depth-enhanced 2D association, some works explore explicit 3D geometric reasoning or Bird's-Eye-View (BEV) representations for tracking, such as GRASPTrack~\cite{han2025grasptrack} and DepthTrack~\cite{tran2025depthtrack}. 

However, most depth-informed MOT methods still use depth as an auxiliary detection-level, frame-level, or short-term association cue~\cite{wang2025pd,liu2025sparsetrack,cui2026depthsort,wu2024depthmot,deng2025deptrmot}, rather than modeling the long-term evolution of instance depth along trajectories. This limits their ability to exploit persistent spatial dynamics and may weaken association robustness under fluctuating monocular depth predictions.

%% file: body/method.tex
\section{Method}
We formulate online panoramic multi-object tracking as a tracking-by-detection problem that jointly accounts for the periodic topology of equirectangular observations and the temporal evolution of geometric cues. 
Our central premise is that horizontal target motion should not be modeled solely in bounded Cartesian image coordinates, and instance depth should not be used merely as an independent frame-wise observation. Instead, horizontal motion is lifted into a continuous cylindrical angular state, while depth is maintained as a temporally evolving trajectory-level state. 
Based on this formulation, we present \textbf{CylindTrack}, a topology-consistent tracking-by-detection framework that addresses two key challenges in panoramic MOT: depth-temporal trajectory modeling and cylindrical topological continuity.

\textbf{Depth-Temporal Trajectory Modeling (DTM)} is proposed as a temporally filtered trajectory state, to provide stable geometric constraints for identity association under occlusion and spatial ambiguity (Section~\ref{subsec:temporal-aware-depth-modeling}). To improve the reliability of depth observations, \textbf{Spherical Spatio-Temporal Consistency Learning (SSTC)} combines a query-based Temporal Mixer with Spherical Geometry-aware Attention (SGA), enhancing temporal coherence and spherical geometry consistency (Section~\ref{subsec:spatial-consistency-omnidirectional}).

Panoramic MOT also suffers from a 2D topological mismatch caused by the cyclic horizontal structure of equirectangular images, as discussed in Section~\ref{subsubsec:cartesian-kalman-failure}. To mitigate this issue, \textbf{Topology-Aware Cylindrical Motion Model (TCMM)} is designed to represent horizontal motion as a periodic angular state, avoiding coordinate discontinuities at the left-right image seam and preserving motion continuity across the $(0^\circ/360^\circ)$ boundary (Section~\ref{subsec:cylindrical-angular-representation}).

\subsection{Depth-Temporal Trajectory Modeling}
\label{subsec:temporal-aware-depth-modeling}
Depth provides an important spatial cue for multi-object tracking, especially under occlusion and close target interactions. Conventional MOT detectors represent each detection \(\mathcal{O}_t^i\) by a 2D bounding box \(\mathbf{b}_t^i\), a confidence score \(s_t^i\), and an identity feature \(\mathbf{e}_t^i\).

Recent depth-aware MOT methods further augment this representation with instance-level depth:
\begin{equation}
\mathcal{O}_t^i = \{\mathbf{b}_t^i, s_t^i, \mathbf{e}_t^i, d_t^i\},
\end{equation}
where $d_t^i$ denotes the estimated depth of target $i$ at frame $t$. This formulation enables the tracker to exploit not only image-plane motion and appearance similarity, but also target-specific spatial ordering for trajectory association.
\input{figures/depth_adjacent_eval}

During association, existing depth-informed trackers commonly incorporate depth as an additional~\cite{deng2025deptrmot} pairwise consistency term:
\begin{equation}
\label{eq:cost}
\mathcal{C}_{ij}
=
\lambda_m \mathcal{C}_{\mathrm{motion}}^{ij}
+
\lambda_a \mathcal{C}_{\mathrm{app}}^{ij}
+
\lambda_z |d_{t-1}^i - d_t^j|.
\end{equation}
Although effective, this formulation enforces depth consistency only between adjacent observations, while ignoring the temporal evolution of depth over the entire trajectory. 
As a result, the association cost can be sensitive to monocular depth scale fluctuations and unstable frame-wise predictions.

To verify this limitation, we conduct a depth-only adjacent-frame association experiment on the QuadTrack training set, as shown in Algorithm~\ref{alg:depth_adjacent_eval}. 
Instance depths are pooled from zero-shot depth maps within ground-truth boxes, and adjacent-frame targets are matched by depth difference using Hungarian assignment.
We report adjacent-frame matching accuracy and sequence-level IDF1 to distinguish local matching reliability from accumulated identity consistency.
The diagnostic uses the $17$ QuadTrack training sequences and overlapping two-frame windows with length $W=2$ and stride $s=1$. For each depth estimator, the scalar instance depth is the median of valid depth values inside the SAM2-Tiny mask prompted by the ground-truth box; SAM2-Tiny is used only for this offline diagnostic and is not part of the CylindTrack inference pipeline. 
The recursive branch uses ground-truth identities only to form window-level depth observations, so it is a label-assisted offline diagnostic rather than the annotation-free deployment procedure.
\input{figures/depth_model_accuracy}

As shown in Fig.~\ref{fig:accuracy-comparison}, the relatively low adjacent-frame matching accuracy indicates that raw frame-wise depth predictions remain sensitive to short-term scale fluctuations. In contrast, the significantly higher sequence-level IDF1 scores obtained with the same DTM-based depth filtering strategy demonstrate the importance of temporal depth modeling.
By reducing local depth noise and enforcing trajectory-level consistency, temporal filtering substantially improves identity preservation. 
Moreover, Video Depth Anything achieves the best adjacent-frame matching performance among all evaluated depth models.
Its superiority further corroborates that temporal consistency is crucial for producing reliable depth cues and effective trajectory association in MOT. 
These findings suggest that depth cues are most effective for MOT when modeled as temporally consistent trajectory signals rather than isolated frame-level observations. 
The contribution of DTM is further examined through the ablations in Tables~\ref{tab:ablation} and~\ref{tab:ablation_depth_module}.

Motivated by this observation, we argue that MOT requires temporally consistent depth trajectories rather than isolated frame-wise depth observations, so that depth information can provide stable spatial constraints for resolving occlusion and identity ambiguity. Accordingly, for each active trajectory, we maintain depth as a trajectory-level latent state:
\begin{equation}
\mathbf{x}_t^i =
\begin{bmatrix}
d_t^i \\
\dot{d}_t^i
\end{bmatrix},
\end{equation}
where $d_t^i$ and $\dot{d}_t^i$ denote the depth and its temporal velocity,
respectively. DTM models the depth motion using a 1D constant-velocity Kalman
filter. The state transition is formulated as
\begin{equation}
\mathbf{x}_t^i=\mathbf{F}_{d}\mathbf{x}_{t-1}^i+\mathbf{w}_t,
\qquad
\mathbf{F}_{d}=\begin{bmatrix}1&1\\0&1\end{bmatrix}.
\end{equation}
where $\mathbf{w}_{t}\sim\mathcal{N}(\mathbf{0},\mathbf{Q})$ denotes the process noise, and $\mathbf{Q}$ is the process noise covariance matrix that accounts for the uncertainty of the constant-velocity depth motion assumption.
The predicted state and covariance are then given by
\begin{equation}
\widehat{\mathbf{x}}_t^i=\mathbf{F}_{d}\mathbf{x}_{t-1}^i,
\qquad
\widehat{\mathbf{P}}_t^i=
\mathbf{F}_{d}\mathbf{P}_{t-1}^i\mathbf{F}_{d}^{\top}+\mathbf{Q}.
\end{equation}

The observation model is defined as
\begin{equation}
z_t = \mathbf{H}\mathbf{x}_t + \eta_t,
\quad
\mathbf{H}=[1,0],
\end{equation}
where $z_t$ is the observed depth of a detection, and $\eta_t\sim\mathcal{N}(0,R)$ denotes the observation noise. 
Given the observed depth $z_t^j$ of a candidate detection, we define the trajectory-to-detection depth association cost as
\begin{equation}
\mathcal{C}_{ij}^{\mathrm{dep}}
=
\left|
\hat{d}_{t}^{i}
-
z_t^j
\right|,
\end{equation}
where $\hat{d}_{t}^{i}$ is the predicted depth of trajectory $i$ at frame $t$. Different from the frame-to-frame comparison in Eq.~\eqref{eq:cost}, which directly uses the previous-frame depth $d_{t-1}^i$, our formulation compares the detection depth with a temporally smoothed trajectory prediction. Since $\hat{d}_t^i$ is inferred from historical depth states, it provides a more stable geometric constraint for robust identity association.
\input{figures/detector}
\input{tables/main_QuaTrack_table}
\subsection{Spherical Spatio-Temporal Consistency Learning}
\label{subsec:spatial-consistency-omnidirectional}
Motivated by the temporal instability and geometric misalignment of frame-wise depth cues in panoramic videos, we introduce Spherical Spatio-Temporal Consistency (SSTC) learning to refine depth-aware query representations. SSTC comprises two complementary components: a query-based Temporal Mixer for local temporal scale alignment and Spherical Geometry-aware Attention (SGA) for aligning depth representations with omnidirectional imaging geometry.

During training, following DepTR-MOT, we use the frozen Video Depth Anything-Small model~\cite{video_depth_anything} to generate offline depth supervision for each video clip. This supervision branch provides depth labels and features for alignment with the detector representations, remains frozen throughout training, and is removed at inference. The detector backbone, encoder, prediction decoders, Temporal Mixer, and SGA constitute the trainable inference path. At inference, the Temporal Mixer operates on a window containing only the current frame and preceding frames. Predictions are produced for the final frame of the window, without accessing future observations. As illustrated in Fig.~\ref{fig:main_detector}, the temporally refined query representations are forwarded to the bounding-box and depth prediction branches. The bounding-box decoder produces object locations and the associated query anchors, whereas the depth decoder produces instance-level depth queries that are further refined by SGA before depth prediction.

Given a batch of $B$ video clips, each containing $T$ frames, the input depth queries are denoted by $\mathbf{X}\in\mathbb{R}^{(BT)\times Q\times C}$, where $Q$ is the number of queries and $C$ is the channel dimension. We rearrange the queries along the temporal dimension using $\mathcal{P}(\cdot)$ and apply layer normalization over the temporal axis:
\begin{equation}
\widetilde{\mathbf{X}}
=
\operatorname{LN}_{T}
\left(
\mathcal{P}(\mathbf{X})
\right),
\quad
\mathcal{P}(\mathbf{X})
\in
\mathbb{R}^{B\times Q\times C\times T}.
\end{equation}
The query-based Temporal Mixer (T-Mixer) is formulated as
\begin{equation}
\mathcal{M}_{t}(\widetilde{\mathbf{X}})
=
\operatorname{Conv}_{2}
\left(
\operatorname{GELU}
\left(
\operatorname{Conv}_{1}
\left(
\operatorname{DWConv}_{t}
\left(
\widetilde{\mathbf{X}}
\right)
\right)
\right)
\right),
\end{equation}
\begin{equation}
\mathbf{X}_{\mathrm{tm}}
=
\mathbf{X}
+
\rho\,
\mathcal{P}^{-1}
\left(
\mathcal{M}_{t}(\widetilde{\mathbf{X}})
\right),
\end{equation}
where $\operatorname{DWConv}_{t}$ independently aggregates local temporal context for each query token, and $\operatorname{Conv}_{1}$ and $\operatorname{Conv}_{2}$ perform channel mixing before and after the GELU activation, respectively. The learnable residual scaling factor $\rho$ is initialized to zero, allowing the temporal branch to be progressively introduced during optimization. The temporally refined queries are subsequently processed by $L$ query-wise MLP layers with hidden dimension $H$:
\begin{equation}
\mathbf{Y}
=
\operatorname{MLP}^{(L)}
\circ\cdots\circ
\operatorname{MLP}^{(1)}
\left(
\mathbf{X}_{\mathrm{tm}}
\right),
\quad
\mathbf{Y}
\in
\mathbb{R}^{(BT)\times Q\times C}.
\end{equation}

By operating on depth-query representations within each video batch, the Temporal Mixer aligns their local temporal feature scales and suppresses short-term fluctuations in frame-wise depth estimation. This representation-level refinement does not require explicit object correspondences across frames.
Moreover, because temporal aggregation is performed on the compact set of $Q$ depth queries rather than on $N_m$ dense memory tokens, where $Q\ll N_m$, it introduces substantially lower token-wise computational overhead.

Following DA$^{2}$~\cite{li2025da2depthdirection}, we construct a deterministic spherical geometry prior from the equirectangular camera geometry. Specifically, for each feature scale, pixel centers are mapped to spherical coordinates according to the horizontal and vertical fields of view, and then encoded with sinusoidal Fourier features to obtain content-independent geometry tokens. The resulting multi-scale geometry context is denoted as
\begin{equation}
\mathbf{S}_{\mathrm{geo}}\in\mathbb{R}^{N_s\times C},
\quad
N_s=\sum_{m=1}^{M}H_mW_m .
\end{equation}
Here, $M$ denotes the number of feature scales, $H_m$ and $W_m$ are the height and width of the $m$-th feature map, $C$ is the channel dimension, and $N_s$ is the total number of geometry tokens across all scales. Since $\mathbf{S}_{\mathrm{geo}}$ only depends on the image resolution and camera field of view, it can be reused across samples with the same panoramic geometry.

Different from DA$^{2}$, which injects spherical geometry into image features, we adapt the spherical prior to instance-level depth prediction refinement. Given temporally enhanced depth queries $\mathbf{Y}$, SGA performs cross-attention from depth queries to the geometry context:
\begin{equation}
\mathbf{Q}
=
\operatorname{LN}_{q}(\mathbf{Y})\mathbf{W}^{Q},
\end{equation}
\begin{equation}
[\mathbf{K}_{\mathrm{geo}},\mathbf{V}_{\mathrm{geo}}]
=
\operatorname{LN}_{\mathrm{ctx}}(\mathbf{S}_{\mathrm{geo}})\mathbf{W}^{KV}.
\end{equation}
Here, $\mathbf{Q}$ is the query projection from the temporally enhanced depth queries, while $\mathbf{K}_{\mathrm{geo}}$ and $\mathbf{V}_{\mathrm{geo}}$ are the key and value projections from the spherical geometry context.
To further reduce the cost of geometry-aware interaction, we employ a width-height separated attention strategy. Instead of attending to the full two-dimensional spherical context at once, SGA decomposes the geometry attention into horizontal and vertical branches:
\begin{equation}
\mathbf{A}_{w}
=
\operatorname{Attn}
\left(
\mathbf{Q},
\mathbf{K}_{w},
\mathbf{V}_{w}
\right),
\end{equation}
\begin{equation}
\mathbf{A}_{h}
=
\operatorname{Attn}
\left(
\mathbf{Q},
\mathbf{K}_{h},
\mathbf{V}_{h}
\right).
\end{equation}
\begin{equation}
\mathbf{X}^{\prime}
=\mathbf{Y}
+
\operatorname{Fuse}
\left(
\mathbf{A}_{w},
\mathbf{A}_{h}
\right),
\end{equation}
Here, $\mathbf{K}_{w},\mathbf{V}_{w}$ and $\mathbf{K}_{h},\mathbf{V}_{h}$ denote the geometry keys and values used in the width and height branches, respectively. The fusion operator $\mathrm{Fuse}(\cdot)$ combines the two directional geometry-aware responses. This separated design preserves spherical priors while reducing attention complexity.
For clarity, Fig.~\ref{fig:main_detector} abstracts the horizontal and vertical attention branches and their fusion as a single CrossAttn block.

Fig.~\ref{fig:main_detector} presents the complete architecture of our enhanced detector, which augments the detector with temporal query refinement and spherical geometry-aware modeling to improve the temporal stability and geometric consistency of instance-level depth predictions.

\subsection{Topology-Aware Cylindrical Motion Model}
\subsubsection{Failure of Cartesian Kalman Filtering in Panoramic MOT}
\label{subsubsec:cartesian-kalman-failure}
In panoramic images, the horizontal coordinate is periodic: the left boundary $x\approx0$ and right boundary $x\approx W$ are adjacent in the real scene but far apart in Cartesian image coordinates. When an object crosses the seam, its displacement may be incorrectly computed as
\begin{equation}
\Delta x = x_t - x_{t-1} \approx -W,
\end{equation}
although the true motion is continuous. This error affects velocity estimation, Kalman prediction, IoU matching, and association costs.

The problem is more severe for objects spanning the panorama boundary, where one physical target may appear as two truncated boxes on opposite image sides. Cartesian modeling treats them as distant objects, leading to incorrect box centers, overlap estimates, motion directions, trajectory fragmentation, and ID switches. Therefore, panoramic MOT requires a cylindrical angular representation that preserves horizontal continuity during prediction and association. Our TCMM addresses this issue by converting horizontal positions into an unwrapped radian state, where longitude evolves continuously in the full state vector and is projected back to $[-\pi,\pi]$ only for ERP-domain mapping.

\subsubsection{Cylindrical Angular Representation}
\label{subsec:cylindrical-angular-representation}
We formulate panoramic MOT as an online tracking-by-detection problem. Given frame-wise detections, TCMM replaces Cartesian horizontal motion with a cylindrical angular representation while keeping the vertical coordinate in the pixel domain. This design preserves horizontal periodicity and enables topology-aware association near panoramic boundaries. During online association, predicted trajectories and detections are matched using horizontal-periodic overlap, trajectory-level depth consistency, angular coherence, and detection confidence.

To integrate it into a standard online TBD pipeline, we retain the conventional detection and trajectory definitions while reformulating their motion states and association measures in cylindrical space. Let the video sequence and frame-$t$ detections be
\begin{equation}
\mathcal{I}=\{I_t\}_{t=1}^{T}, 
\
\mathcal{D}_t=\{o_{t,j}\}_{j=1}^{M_t},
\
o_{t,j}=(b_{t,j},s_{t,j},d_{t,j}),
\end{equation}
where \(b_{t,j}=(x_{t,j},y_{t,j},w_{t,j},h_{t,j})\) is the bounding box, \(s_{t,j}\) is the confidence score, and \(d_{t,j}\) is the depth measurement. The tracker estimates a set of trajectories
\begin{equation}
\hat{\mathcal{Y}}
=
\left\{
\{(t,\hat{b}_{k,t},k)\}_{t\in\mathcal{T}_k}
\right\}_{k=1}^{K},
\end{equation}
where \(k\) denotes the trajectory identity and \(\mathcal{T}_k\) is the set of frame indices.
For motion prediction, each trajectory is represented in a cylindrical state space:
\begin{equation}
\mathbf{m}_t =
(\theta_t, y_t, a_t, h_t, \dot{\theta}_t, \dot{y}_t, \dot{a}_t, \dot{h}_t)^{\top},
\end{equation}
where \(\theta_t\) is the horizontal angular coordinate, \(y_t\) is the vertical image coordinate, \(a_t\) is the aspect ratio, and \(h_t\) is the box height. Given the horizontal box center \(x_c=x+w/2\) and image width \(W\), the angular coordinate is computed as
\begin{equation}
\theta =
2\pi
\left(
\frac{x_c}{W}
-
\frac{1}{2}
\right).
\end{equation}

The state follows a constant-velocity transition:
\begin{equation}
\mathbf{m}_{t\mid t-1}=\mathbf{F}_{c}\mathbf{m}_{t-1\mid t-1}.
\end{equation}
Here, $\mathbf{F}_{c}$ is the transition matrix for the eight-dimensional cylindrical motion state:
\[
\mathbf{F}_{c}=
\begin{bmatrix}
\mathbf{I}_{4}&\mathbf{I}_{4}\\
\mathbf{0}_{4}&\mathbf{I}_{4}
\end{bmatrix},
\]
where one frame is used as the unit time step.

To avoid discontinuous updates across the panorama boundary, an observed angle \(\theta\) is converted to its nearest equivalent form with respect to the predicted angle \(\hat{\theta}\):
\begin{equation}
\theta^{\star}
=
\hat{\theta}
+
\operatorname{wrap}
\left(
\theta-\hat{\theta}
\right),
\end{equation}
\begin{equation}
\mathrm{wrap}(\delta)=((\delta+\pi)\bmod 2\pi)-\pi
\end{equation}
where \(\operatorname{wrap}(\cdot)\) maps an angular difference to \([-\pi,\pi)\). The Kalman update is then performed using \(\theta^{\star}\), preserving motion continuity across the \(0^\circ/360^\circ\) boundary. For completeness, each trajectory is also associated with a one-dimensional depth state \(\mathbf{r}_t=(d_t,\dot{d}_t)^\top\), which is maintained by the temporal-aware depth model in Section~\ref{subsec:temporal-aware-depth-modeling}.
We maintain the longitude state in the unwrapped lifted angular space \(\mathbb{R}\), allowing angles to evolve continuously across the \(0^\circ/360^\circ\) seam. Thus, standard Kalman state and covariance updates are preserved, and wrapping is only used for observation lifting and projection back to the canonical ERP image domain.

To measure spatial overlap under horizontal periodicity, we define a cylindrical overlap representation. Given a bounding box \(b=(x,y,w,h)\), its center is
\begin{equation}
x_c=x+\frac{w}{2},
\qquad
y_c=y+\frac{h}{2}.
\end{equation}
The horizontal center and width are converted into angular form:
\begin{equation}
\theta=2\pi\left(\frac{x_c}{W}-\frac{1}{2}\right),
\qquad
\Delta\theta=\frac{2\pi w}{W}.
\end{equation}
We represent the box as
\begin{equation}
R(b)=(\theta,y_c,\Delta\theta,h),
\end{equation}
where the horizontal interval is evaluated on a circular domain:
\begin{equation}
I_{\theta}(b)
=
\left[
\theta-\frac{\Delta\theta}{2},
\theta+\frac{\Delta\theta}{2}
\right]_{2\pi},
\end{equation}
and the vertical interval remains in the image pixel domain:
\begin{equation}
I_y(b)
=
\left[
y_c-\frac{h}{2},
y_c+\frac{h}{2}
\right].
\end{equation}
This representation preserves left--right boundary continuity while avoiding unnecessary latitude-area correction along the vertical direction.
For two boxes \(b_i\) and \(b_j\), the horizontal periodic overlap is
\begin{equation}
\ell_{\theta}(i,j)
=
\left|
I_{\theta}(b_i)
\cap
I_{\theta}(b_j)
\right|_{2\pi}.
\end{equation}
The vertical overlap is
\begin{equation}
\ell_y(i,j)
=
\max\left(
0,\,
\min\left(y_i^{+},y_j^{+}\right)
-
\max\left(y_i^{-},y_j^{-}\right)
\right),
\end{equation}
where \(y_i^{-}=y_{c,i}-h_i/2\) and \(y_i^{+}=y_{c,i}+h_i/2\). The cylindrical intersection and box measures are defined as
\begin{equation}
A_{\cap,\mathrm{hp}}(i,j)=\ell_{\theta}(i,j)\ell_y(i,j),
\qquad
A_{\mathrm{hp}}(i)=\Delta\theta_i h_i.
\label{eq:cylindrical_intersection}
\end{equation}
The horizontal-periodic pixel-vertical IoU (HPV-IoU) is computed as
\begin{equation}
\operatorname{IoU}_{\mathrm{hp}}(i,j)
=
\frac{
A_{\cap,\mathrm{hp}}(i,j)
}{
A_{\mathrm{hp}}(i)
+
A_{\mathrm{hp}}(j)
-
A_{\cap,\mathrm{hp}}(i,j)
+
\epsilon
}.
\end{equation}
The corresponding overlap cost is
\begin{equation}
C_{\mathrm{hp}}(i,j)
=
1-
\operatorname{IoU}_{\mathrm{hp}}(i,j).
\label{eq:overlap_cost}
\end{equation}
In addition to topology-aware spatial overlap, we introduce a normalized depth consistency term. For a predicted trajectory \(i\) and detection \(j\), the normalized depth cost is
\begin{equation}
C_z(i,j)
=
\frac{
|\hat{d}_i-z_j|
}{
\max_{p,q}|\hat{d}_p-z_q|+\epsilon
},
\end{equation}
where \(\hat{d}_i\) is the trajectory-level depth prediction from DTM and \(z_j\) is the observed depth of detection \(j\). The depth-aware cylindrical cost is
\begin{equation}
C_{d\text{-}\mathrm{hp}}(i,j)
=
(1-\lambda_z)C_{\mathrm{hp}}(i,j)
+
\lambda_z C_z(i,j).
\label{eq:depth_assoc_cost}
\end{equation}
We further impose angular consistency between predicted trajectories and detections:
\begin{equation}
C_{\theta}(i,j)
=
\frac{
\left|
\operatorname{wrap}(\theta_i-\theta_j)
\right|
}{\pi}.
\end{equation}
The first-stage association cost $\tilde{C}(i,j)$, fused with detection confidence, is defined as
\begin{equation}
C_{\mathrm{assoc}}(i,j)
=
(1-\lambda_{\theta})C_{d\text{-}\mathrm{hp}}(i,j)
+
\lambda_{\theta}C_{\theta}(i,j),
\label{eq:assoc_cost_params}
\end{equation}
\begin{equation}
\tilde{C}(i,j)
=
1-
\left(
1-C_{\mathrm{assoc}}(i,j)
\right)s_j,
\label{eq:assoc_cost}
\end{equation}
where \(\lambda_{\theta}\) controls the contribution of angular consistency.

At each frame, association is solved by constrained linear assignment:
\begin{equation}
X_t^{*}
=
\arg\min_{
X\in\{0,1\}^{|\mathcal{T}_t|\times|\mathcal{Z}_t|}
}
\sum_{i,j}
X_{ij}\tilde{C}(i,j).
\end{equation}
\begin{equation}
\begin{aligned}
\mathrm{s.t.}\quad
&\sum_j X_{ij}\leq 1,\ \forall i,\quad
\sum_i X_{ij}\leq 1,\ \forall j,\\
&\sum_{i,j}X_{ij}
=\min(|\mathcal T_t|,|\mathcal Z_t|).
\end{aligned}
\label{eq:assignment_constraint}
\end{equation}

The assignment enforces one-to-one matching with maximum cardinality, while invalid matches are subsequently rejected according to the matching threshold.
The accepted first-stage matches are
\begin{equation}
\mathcal{M}_t^{(1)}
=
\left\{
(i,j)\mid
X_{ij}^{*}=1,\ 
\tilde{C}(i,j)\le \tau_m
\right\}.
\end{equation}
During online inference, detections are divided into high- and low-confidence sets according to thresholds \(\tau_h\) and \(\tau_l\). High-confidence detections are first matched to predicted trajectories using the full association cost in Eq.~\eqref{eq:assoc_cost}. Unmatched active trajectories are then associated with low-confidence detections using cylindrical IoU alone, which helps recover targets with temporarily reduced detection confidence. Unmatched high-confidence detections above the initialization threshold \(\tau_{\mathrm{init}}\) are initialized as new trajectories, while unmatched tracks are retained for at most \(B_{\mathrm{lost}}\) frames before removal. Finally, near-duplicate trajectories are suppressed according to cylindrical overlap, retaining the longer-lived trajectory.

As shown in Fig.~\ref{fig:lifting}, TCMM lifts ERP detections into a depth-aware cylindrical state space, where horizontal coordinates are represented as periodic longitudes and depth provides radial geometric constraints for topology-consistent association.

%% file: figures/depth_adjacent_eval.tex
\begin{algorithm}[!t]
\footnotesize
\setstretch{1.08}
\caption{Depth-Based Trajectory Association}
\label{alg:depth_adjacent_eval}

\KwIn{
Video sequence $\{I_t\}_{t=1}^{T}$, ground-truth boxes
$\{\mathcal{B}_t\}_{t=1}^{T}$,
pretrained depth model $M$
}
\KwOut{
IDF1 and adjacent-frame matching accuracy $\mathrm{Acc}$
}

\BlankLine
$\mathcal{W} \leftarrow
\mathrm{SlidingWindow}(\{I_t\}_{t=1}^{T}, W=2, s=1)$
\tcp*[r]{Window length $W$ and stride $s$}

$\mathcal{S}_1 \leftarrow \varnothing$\;

\BlankLine
\ForEach{window $w_t=(I_t,I_{t+1}) \in \mathcal{W}$}{
    Load frames $I_t$ and $I_{t+1}$ from $w_t$\;

    \BlankLine
    $\hat{D}_t \leftarrow M(I_t)$
    \tcp*[r]{Depth map at $t$}
    $\hat{D}_{t+1} \leftarrow M(I_{t+1})$
    \tcp*[r]{Depth map at $t+1$}

    \BlankLine
    $\mathcal{G}_t \leftarrow
    \mathrm{ID}(\mathcal{B}_t)
    \cap \mathrm{ID}(\mathcal{B}_{t+1})$
    \tcp*[r]{Shared GT IDs}

    $\mathcal{B}_t^{c} \leftarrow
    \{b_t^i \in \mathcal{B}_t \mid
    \mathrm{ID}(b_t^i) \in \mathcal{G}_t\}$\;

    $\mathcal{B}_{t+1}^{c} \leftarrow
    \{b_{t+1}^j \in \mathcal{B}_{t+1} \mid
    \mathrm{ID}(b_{t+1}^j) \in \mathcal{G}_t\}$\;

    \BlankLine
    \ForEach{target $b_t^i \in \mathcal{B}_t^{c}$}{
        $\hat{d}_t^i \leftarrow
        \mathrm{Pool}(\hat{D}_t(b_t^i))$
        \tcp*[r]{Median valid depth}
    }

    \ForEach{target $b_{t+1}^j \in \mathcal{B}_{t+1}^{c}$}{
        $\hat{d}_{t+1}^j \leftarrow
        \mathrm{Pool}(\hat{D}_{t+1}(b_{t+1}^j))$\;
    }

    $(\mathcal{S}_{t+1}^{-},
      \{\tilde{d}_{t+1}^{i}\}_i)
    \leftarrow
    \mathrm{DTM}(\mathcal{S}_t,
      \{\hat{d}_t^i\}_i,\mathrm{predict})$\;

    \BlankLine
    Construct cost matrices $C^{dep}$ and $C^{DTM}$\;

    \ForEach{$b_t^i \in \mathcal{B}_t^{c}$}{
        \ForEach{$b_{t+1}^j \in \mathcal{B}_{t+1}^{c}$}{
            $c_{ij}^{dep} \leftarrow
            |\hat{d}_t^i-\hat{d}_{t+1}^j|$\;

            $c_{ij}^{DTM} \leftarrow
            |\tilde{d}_{t+1}^{i}-\hat{d}_{t+1}^{j}|$\;
        }
    }

    \BlankLine
    $\pi_t \leftarrow \mathrm{Hungarian}(C^{dep})$\;

    $\mathcal{M}_t \leftarrow
    \{(b_t^i,b_{t+1}^{\pi_t(i)})\}$\;

    $\pi_t^{DTM} \leftarrow
    \mathrm{Hungarian}(C^{DTM})$\;

    $\mathcal{M}_t^{DTM} \leftarrow
    \{(b_t^i,b_{t+1}^{\pi_t^{DTM}(i)})\}$\;

    $\mathcal{S}_{t+1} \leftarrow
    \mathrm{DTM}(\mathcal{S}_{t+1}^{-},
      \pi_t^{DTM},
      \{\hat{d}_{t+1}^{j}\}_j,\mathrm{update})$\;
}

\BlankLine
$(\mathrm{IDF1},\mathrm{Acc}) \leftarrow
\mathrm{EvaluateMetrics}\!\left(
\{\mathcal{M}_t^{DTM}\}_{t=1}^{T-1},
\{\mathcal{M}_t\}_{t=1}^{T-1},
\{\mathcal{B}_t^{c},\mathcal{B}_{t+1}^{c}\}_{t=1}^{T-1}
\right)$\;

\Return $\mathrm{IDF1},\mathrm{Acc}$\;
\end{algorithm}

%% file: figures/depth_model_accuracy.tex
\begin{figure}[!t]
    \centering
    \includegraphics[width=\linewidth]{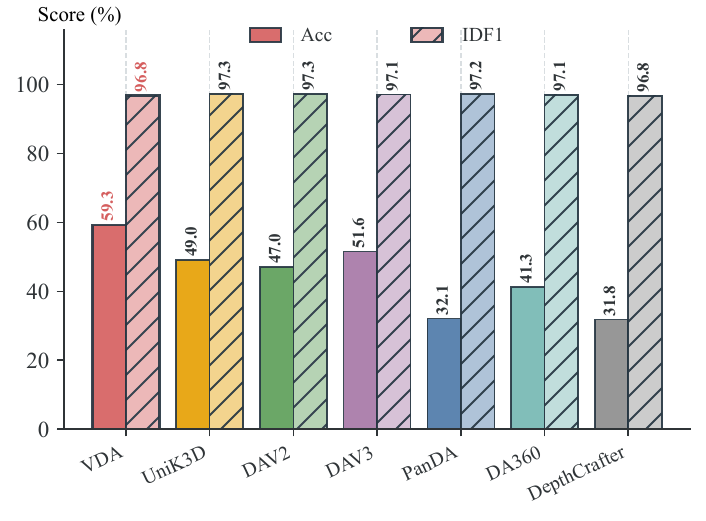}
    \vskip-3ex
    \caption{Performance for depth-only trajectory association across all sequences, measured by temporal association performance (IDF1) and adjacent-frame association accuracy (Acc). Adjacent-frame accuracy is evaluated using raw frame-wise depth differences, whereas sequence-level IDF1 is computed by recursively associating detections; each matched observation updates the trajectory depth state via the DTM filter. We evaluate multiple pretrained zero-shot depth estimation models~\cite{cao2025panda,jiang2025da360,video_depth_anything,piccinelli2025unik3d,depth_anything_v2,depthanything3,hu2025-DepthCrafter}, where all methods are evaluated using their smallest-parameter variant. 
    }
    \label{fig:accuracy-comparison}
\end{figure}

%% file: figures/detector.tex
\begin{figure*}[t]
    \centering
    \includegraphics[
        width=\textwidth,
        trim=60 10 55 40,
        clip
    ]{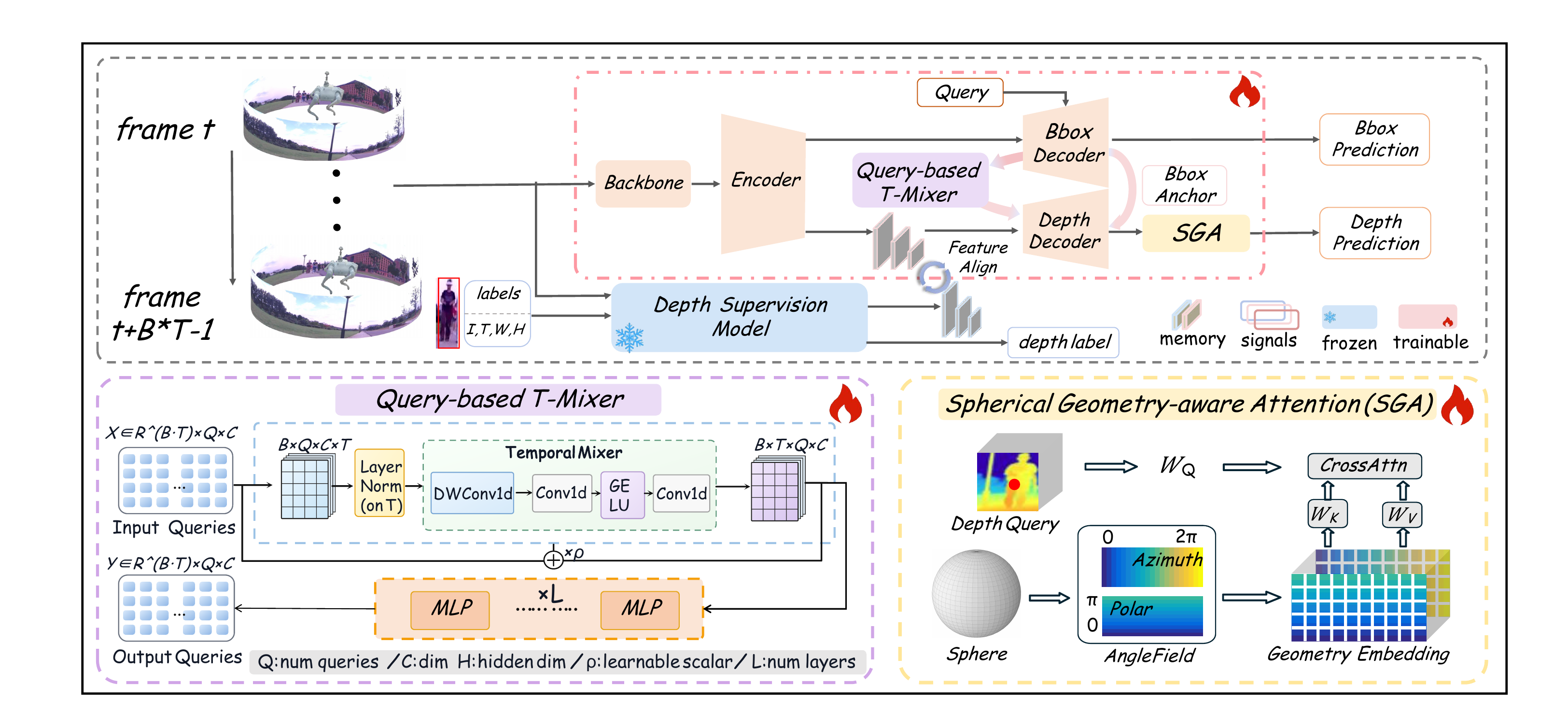}
    \vskip-1ex
    \caption{Overview of the proposed \textbf{Spherical Spatio-Temporal Consistency Learning (SSTC)} method for the depth-enhanced detector. SSTC improves depth-aware instance representations by combining the query-based Temporal Mixer with Spherical Geometry-aware Attention (SGA). 
    The query-based Temporal Mixer performs local temporal scale alignment of depth-query representations within each video batch, reducing frame-to-frame depth fluctuations, while SGA incorporates spherical geometric priors to better align depth prediction with equirectangular panoramic geometry. 
    The refined depth observations are used as reliable geometric cues for subsequent depth-guided trajectory association. The snowflake denotes the frozen depth-supervision branch used only during training, whereas the flame denotes trainable components.
    }
    \label{fig:main_detector}
\end{figure*}

%% file: tables/main_QuaTrack_table.tex
\begin{table*}[t]
\centering
\small
\setlength{\tabcolsep}{4.2pt}
\renewcommand{\arraystretch}{0.8}

\newcommand{\posgain}[1]{\textcolor{green!50!black}{\scriptsize\,($+#1$)}}
\newcommand{\neggain}[1]{\textcolor{gray}{\scriptsize\,($#1$)}}

\resizebox{\textwidth}{!}{
\begin{tabular}{l|c|rrrrrr}
\toprule[1.2pt]
\textbf{Tracker} & \textbf{Detector} 
& \textbf{HOTA$\uparrow$} 
& \textbf{BCIC$\uparrow$} 
& \textbf{AssA$\uparrow$} 
& \textbf{IDF1$\uparrow$} 
& \textbf{MOTA$\uparrow$} 
& \textbf{FPS$\uparrow$} \\
\midrule

\multirow{2}{*}{SORT~\cite{bewley2016sort}}
& DepTR-MOT & 18.195 & 21.114 & 13.951 & 17.382 & 4.745 & 27.88 \\
& \(\mathrm{CylindTrack}_{\mathrm{Det}}\)     
& 20.880\posgain{2.685} 
& 20.371\neggain{-0.743}
& 18.267\posgain{4.316} 
& 21.128\posgain{3.746} 
& 7.664\posgain{2.919} 
& 28.25\posgain{0.37} \\
\midrule

\multirow{2}{*}{DeepSORT~\cite{wojke2017deepsort}}
& DepTR-MOT & 26.030 & 28.202 & 24.242 & 29.060 & -6.267 & 27.12 \\
& \(\mathrm{CylindTrack}_{\mathrm{Det}}\)     
& 26.400\posgain{0.370} 
& 27.957\neggain{-0.245} 
& 24.593\posgain{0.351} 
& 30.633\posgain{1.573} 
& 2.034\posgain{8.301} 
& 27.81\posgain{0.69} \\
\midrule

\multirow{2}{*}{Bot-SORT~\cite{aharon2022botsortrobustassociationsmultipedestrian}}
& DepTR-MOT & 15.023 & 25.343 & 14.601 & 12.323 & -148.390 & 29.24 \\
& \(\mathrm{CylindTrack}_{\mathrm{Det}}\)     
& 16.985\posgain{1.962} 
& 24.043\neggain{-1.300} 
& 15.149\posgain{0.548} 
& 15.345\posgain{3.022} 
& -83.617\posgain{64.773} 
& \textbf{29.65}\posgain{0.41} \\
\midrule

\multirow{2}{*}{OC-SORT~\cite{cao2023observation}}
& DepTR-MOT & 18.376 & 19.446 & 15.829 & 18.978 & 3.550 & 23.88 \\
& \(\mathrm{CylindTrack}_{\mathrm{Det}}\)     
& 20.705\posgain{2.329} 
& 19.802\posgain{0.356} 
& 19.708\posgain{3.879} 
& 22.575\posgain{3.597} 
& 5.382\posgain{1.832} 
& 25.29\posgain{1.41} \\
\midrule

\multirow{2}{*}{HybridSORT~\cite{yang2024hybrid}}
& DepTR-MOT & 20.293 & 20.829 & 18.366 & 21.114 & 3.066 & 23.62 \\
& \(\mathrm{CylindTrack}_{\mathrm{Det}}\)     
& 20.813\posgain{0.520} 
& 22.079\posgain{1.250} 
& 19.393\posgain{1.027} 
& 22.090\posgain{0.976} 
& 4.633\posgain{1.567} 
& 25.05\posgain{1.43} \\
\midrule

\multirow{2}{*}{DiffMOT~\cite{lv2024diffmot}}
& DepTR-MOT & 13.058 & 28.530 & 12.806 & 10.785 & -199.180 & 29.05 \\
& \(\mathrm{CylindTrack}_{\mathrm{Det}}\)     
& 15.782\posgain{2.724} 
& 26.131\neggain{-2.399} 
& 15.417\posgain{2.611} 
& 14.346\posgain{3.561} 
& -124.650\posgain{74.530} 
& 29.42\posgain{0.37} \\
\midrule

\multirow{2}{*}{ByteTrack~\cite{zhang2022bytetrack}}
& DepTR-MOT 
& 31.026 
& 30.578
& 29.786 
& 34.109 
& 14.387
& 29.09 \\
& \(\mathrm{CylindTrack}_{\mathrm{Det}}\)     
& 30.009\neggain{-1.017} 
& 29.317\neggain{-1.261} 
& 27.701\neggain{-2.085} 
& 34.641\posgain{0.532} 
& 20.046\posgain{5.659} 
& 29.21\posgain{0.12} \\
\midrule[\heavyrulewidth]
\multicolumn{2}{c|}{\(\mathrm{OmniTrack}_{\mathrm{TBD}}\)~\cite{luo2025omnitrack}} 
& 19.563 & 22.189 & 19.836 & 16.769 & -13.885
& 24.56 \\

\midrule			
\multicolumn{2}{c|}{\(\mathrm{CylindTrack}_{\mathrm{TBD}}\)} 
& \textbf{33.674} & \textbf{54.957} & \textbf{34.665} & \textbf{40.446} & \textbf{20.583}
& 28.56 \\

\bottomrule[1.2pt]
\end{tabular}
}
\vskip-1ex
\caption{Main results under the depth-enabled setting on QuadTrack~\cite{luo2025omnitrack}. 
Parenthesized colored values denote the gain of \(\mathrm{CylindTrack}_{\mathrm{Det}}\) over DepTR-MOT~\cite{deng2025deptrmot}: green for improvement and gray for degradation. 
Only column-best results are bolded. 
All metrics are higher-is-better. BCIC denotes Boundary Crossing Identity Consistency, a boundary-aware metric defined in Section~\ref{sec:Metrics}. The performance gains come from both improved depth-aware detection in \(\mathrm{CylindTrack}_{\mathrm{Det}}\) and enhanced topology-aware association in \(\mathrm{CylindTrack}_{\mathrm{TBD}}\).}
\vskip-1ex
\label{tab:main-depth-results}
\end{table*}

%% file: body/experiment.tex
\input{figures/lift_to_3d}

\input{tables/main_JRDB_table}
\section{Experiments}

\subsection{Experiment Setup}
\subsubsection{Datasets}
To comprehensively evaluate our approach, we conduct experiments on panoramic tracking datasets JRDB~\cite{martin2021jrdb} and QuadTrack~\cite{luo2025omnitrack}.

\textbf{QuadTrack.}
QuadTrack is a panoramic multi-object tracking dataset captured by a quadruped robot with a \(360^\circ\) visual sensor.
It contains $32$ sequences, split into training and evaluation subsets.
The dataset follows the MOT-style annotation format and includes target identities and bounding boxes across video frames.
Due to quadrupedal locomotion and omnidirectional imaging, it exhibits strong ego-motion, vertical camera shake, motion blur, illumination variation, and projection distortion, making it a challenging benchmark for embodied panoramic tracking.

\textbf{JRDB.}
JRDB is a robotic perception dataset collected by a mobile robot in complex indoor and outdoor human-centered environments.
It contains $54$ sequences, with $20$ for training, $7$ for validation, and $27$ for testing under the official split.
Its panoramic visual observations involve crowded scenes, frequent occlusions, small-scale targets, appearance ambiguity, and fast relative motion.
These challenges make JRDB suitable for evaluating the robustness and identity consistency of MOT methods in real-world panoramic robotic scenarios.

\subsubsection{Metrics}
\label{sec:Metrics}
Following prior panoramic MOT studies, such as OmniTrack~\cite{luo2025omnitrack} and DepTR-MOT~\cite{deng2025deptrmot}, we evaluate the proposed method using HOTA, MOTA, IDF1, AssA, DetA, and FPS~\cite{luiten2021hota,bernardin2008clearmot,ristani2016idmetrics,schuhmacher2008consistent}. We place particular emphasis on identity-related metrics, including IDF1 and AssA, to assess trajectory continuity.
To evaluate identity preservation during boundary transitions, we introduce
\textbf{Boundary Crossing Identity Consistency (BCIC)}, a boundary-aware
association metric for multi-object tracking. Our BCIC metric is motivated by IDF1~\cite{ristani2016idmetrics} and the association accuracy (AssA) in HOTA~\cite{luiten2021hota}. Instead of evaluating identity consistency over complete trajectories, BCIC
measures whether a target preserves the same predicted identity within a local
temporal window centered at a boundary-crossing event.
Given a predefined boundary $B$, a crossing event $E_i^k=(i,t_c)$ is detected
when the ground-truth trajectory of target $i$ moves from one side of $B$ to the
other at frame $t_c$. For each event, we define a temporal evaluation window
around the crossing frame:
\[
W(E_i^k)=[t_c-\Delta,\,t_c+\Delta],
\]
where $\Delta$ denotes the window radius. Let $p_t$ be the predicted identity
associated with target $i$ at frame $t$. The dominant identity within the
window is defined as
\[
p^*=
\arg\max_j
\sum_{t\in W(E_i^k)}
\mathbf{1}(p_t=j).
\]
The identity consistency score for this crossing event is then computed as
\[
IC(E_i^k)=
\frac{
\sum_{t\in W(E_i^k)}
\mathbf{1}(p_t=p^*)
}
{|W(E_i^k)|}.
\]
Finally, BCIC is obtained by averaging over all $M$ boundary-crossing events:
\[
\mathrm{BCIC}
=
\frac{1}{M}
\sum_{i,k}
IC(E_i^k).
\]
A higher BCIC indicates that the tracker maintains more stable identities when
objects cross critical boundaries.

For reproducibility, with ERP width $W_{\mathrm{img}}$ we define the boundary
region as $[0,0.12W_{\mathrm{img}})\cup(0.88W_{\mathrm{img}},W_{\mathrm{img}})$.
Using consecutive available ground-truth observations, a crossing is retained
only when the frame gap is at most $5$, the observations lie in opposite bands,
and the horizontal jump exceeds $0.45W_{\mathrm{img}}$; duplicate events are
removed. The fixed operating points are $\Delta=15$ for QuadTrack (31-frame
windows, 34 events) and $\Delta=45$ for JRDB (91-frame windows, 24 events), and
the event lists are shared by all trackers within each dataset. For each target
and frame, the prediction with maximum planar IoU is selected independently and
accepted only if IoU is at least $0.5$; no global one-to-one assignment is used.
Missed detections, missing annotations, and out-of-range frames remain in the
window denominator and contribute zero. The dominant identity is selected only
from matched IDs; ties use the earliest occurrence, and an event with no match
receives score zero. The implementation returns BCIC in $[0,1]$, whereas the
tables report $100\times\mathrm{BCIC}$.

\subsubsection{Implementation Details}
We use DepTR-MOT~\cite{deng2025deptrmot} as the closest detector-side baseline. Following the baseline configuration, we generate the offline depth supervision using the lightweight Small checkpoint of Video Depth Anything~\cite{video_depth_anything} and obtain instance masks using the SAM2-Tiny checkpoint~\cite{ravi2024sam2}. The input resolution is set to $480 \times 2048$ for QuadTrack~\cite{luo2025omnitrack} and $480 \times 3760$ for JRDB~\cite{martin2021jrdb}, without additional data augmentation. Training is conducted for $5$ epochs on each dataset using AdamW with an initial learning rate of $2.5 \times 10^{-4}$ and a batch size of $1$. For a controlled comparison, $\mathrm{CylindTrack}_{\mathrm{Det}}$ follows the same supervised-training and annotation-free inference protocol as DepTR-MOT, while the backbone, supervision sources, input resolutions, optimization settings, data processing, and evaluation configurations are kept unchanged. 
All experiments are conducted on an NVIDIA GeForce RTX 3090 GPU with PyTorch 2.7.1 and CUDA 12.6.

\input{tables/ablation}

\subsection{Quantitative Comparison}
As shown in Tables~\ref{tab:main-depth-results} and~\ref{tab:depth-detector-comparison}, across the tested trackers, \(\mathrm{CylindTrack}_{\mathrm{Det}}\) improves most identity-related metrics, with the most stable gains observed on IDF1 and AssA, while other metrics and FPS remain tracker-dependent. Across the seven trackers, replacing the DepTR-MOT detector with CylindTrackDet increases the mean HOTA, AssA, and IDF1 scores by $6.74\%$, $8.22\%$, and $11.83\%$, respectively, on QuadTrack, and by $5.42\%$, $11.59\%$, and $6.15\%$, respectively, on JRDB. Mean MOTA increases from $-46.870$ to $-24.073$ on QuadTrack and from $16.777$ to $24.684$ on JRDB, corresponding to gains of $22.797$ and $7.906$ percentage points, respectively. The relative improvements in HOTA, AssA, and IDF1 are computed from the mean scores across the seven trackers. Detector replacement alone changes mean BCIC from $24.863$ to $24.243$ on QuadTrack and from $26.587$ to $27.307$ on JRDB, whereas the complete CylindTrackTBD framework achieves BCIC scores of $54.957$ and $35.365$, respectively.

Compared with the strongest DepTR-MOT+ByteTrack baseline, our full Tracking-by-Detection (TBD) framework improves IDF1 and AssA by 18.58\% and 16.38\% on QuadTrack, and by 10.66\% and 18.74\% on JRDB, respectively, with BCIC gains of 24.38 and 13.51 percentage points. For fairness, we follow the same depth-aware and depth-free association strategies as DepTR-MOT and OmniTrack, while keeping the hyperparameters unchanged within each experimental group.
\input{tables/ablation_5}

Overall, these results validate the effectiveness of \(\mathrm{CylindTrack}_{\mathrm{TBD}}\). Fig.~\ref{fig:lifting} provides an intuitive illustration of the proposed 2D-to-3D lifting process, where 2D bounding boxes and depth cues are consistently transformed into topology-consistent cylindrical state space. This depth-aware cylindrical representation enables effective motion modeling at low computational cost, allowing our method to improve identity-oriented performance on average and achieve stable gains in most tracker settings.

\subsection{Ablation Studies}
In Experiments \eid{2}-\eid{4} of Table~\ref{tab:ablation}, we separately decouple the three components to analyze the contribution of each individual module. 
In Experiments \eid{5}-\eid{7}, we further investigate the coupling effects among different components. 
In Table~\ref{tab:cyl_sph_ablation}, we investigate the contribution of each TCMM component and compare cylindrical and spherical representations. In Table~\ref{tab:ablation_depth_module}, we further analyze the contribution of each depth-related component and their coupling effects.

\subsubsection{Depth-Aware Cylindrical Motion Modeling}
We first conduct an ablation analysis of the main designs in Table~\ref{tab:ablation}.
As shown by Experiments~\eid{1} and~\eid{2}, enabling DTM improves HOTA, IDF1, and AssA by $0.361$, $0.916$, and $0.536$, respectively. 
This suggests that depth cues mainly provide additional identity-discriminative information for association. This benefit is further demonstrated when DTM is combined with TCMM: Experiment~\eid{5} achieves higher HOTA, IDF1, and AssA than Experiment~\eid{4}, indicating that temporally modeled depth cues further enhance association when motion is represented in a topology-consistent panoramic space.

We then analyze the role of Spherical Spatio-Temporal Consistency (SSTC). As shown by Experiments~\eid{4} and~\eid{6}, introducing this strategy together with the TCMM improves HOTA, IDF1, AssA, and MOTA by $1.536$, $3.099$, $2.500$, and $7.014$ points, respectively. 
Compared with the baseline, this combination brings gains of $1.974$ in HOTA, $5.536$ in IDF1, $3.420$ in AssA, and $5.866$ in MOTA. 
These results verify that reliable temporal depth cues can provide stable identity-consistency improvements, especially when the representation respects the periodic topology of panoramic views.
\input{tables/ablation_depth}
For boundary scenarios, the coupling between spatio-temporally consistent depth cues and the TCMM is particularly critical. The TCMM provides a non-Euclidean state formulation for cross-boundary motion, while reliable depth cues offer additional identity separability under occlusion and ambiguous 2D observations. 
Their combination fully exploits depth information for panoramic MOT, as evidenced by Experiment~\eid{8}, where the full model achieves the best overall performance in HOTA, IDF1, AssA, and MOTA.

\input{figures/visualization_boundary}

Finally, Experiment~\eid{3} shows that using the spatio-temporal component alone brings only limited performance changes, with fluctuations in identity-related metrics and no consistent gains across all evaluation measures. Similarly, Experiment~\eid{7}, which combines depth-aware association and spatio-temporal modeling without the TCMM, still fails to achieve clear improvements over the full model. These results suggest that depth or temporal cues alone are insufficient for robust identity preservation under panoramic boundary conditions. This conclusion is also consistent with the observations in Fig.~\ref{fig:pano_challenge} (c) / (d): boundary-crossing regions often suffer from annotation discontinuities caused by the periodic panoramic topology, while occlusions in general scene regions can weaken detector responses and reduce the reliability of 2D-based association. These visual observations further support the experimental finding that coupling depth cues and temporal-geometric consistency with the periodic angular state representation is essential for maintaining cross-boundary identity consistency. Benefiting mainly from the topology-aware periodic state introduced by TCMM, Fig.~\ref{fig:visualization_quadtrack_jrdb} shows that CylindTrack preserves trajectory continuity across ERP boundaries on both QuadTrack and JRDB, whereas direct UV tracking is prone to fragmented associations under periodic spatial transformations, consistent with its 24.38- and 13.51-point BCIC gains over the strongest baseline.
\input{figures/visualization_quadtrack}

Table~\ref{tab:cyl_sph_ablation} reports the ablation results of different periodic geometric modeling components. Compared with the image-plane baseline in Experiment~\eid{1}, using only the cylindrical motion Kalman filter in Experiment~\eid{2} or only the topology-aware IoU in Experiment~\eid{3} brings only marginal changes, indicating that either motion periodicity or overlap periodicity alone is insufficient for robust cross-boundary association. When both components are enabled in Experiment~\eid{4}, the tracker achieves clearer improvements in HOTA, IDF1, and AssA, suggesting that motion prediction and overlap computation should be modeled under a consistent panoramic topology.

The periodic angle cost further contributes to identity-preserving association. Experiment~\eid{5}, which adds the angle cost to the image-plane baseline, improves HOTA, IDF1, and AssA over Experiment~\eid{1}, showing that angular consistency provides useful association cues even without cylindrical motion modeling. More importantly, the full cylindrical setting in Experiment~\eid{6} achieves consistent gains over the baseline, improving HOTA from $30.817$ to $33.659$, IDF1 from $35.498$ to $40.412$, and AssA from $29.255$ to $34.622$.

\subsubsection{Comparison between Cylindrical and Spherical Representations}

The ablation results compare the horizontal-periodic pixel-vertical, cylindrical, and spherical overlap representations within the same association framework in Table~\ref{tab:cyl_sph_ablation}. All variants use the same detections, confidence scores, association settings, and evaluation protocol, ensuring a fair comparison focused solely on the geometric representation.

Motivated by recent spherical state-space modeling for panoramic tracking~\cite{liu2026s3kf}, we do not assume that cylindrical modeling is universally more general than spherical modeling. 
Instead, our association cost is modular with respect to the overlap term, so horizontal-periodic pixel-vertical, cylindrical, and spherical representations can be exchanged without changing the remaining association pipeline.

In the main setting, we use the horizontal-periodic pixel-vertical IoU as shown in Eq.~\ref{eq:cylindrical_intersection}--Eq.~\ref{eq:overlap_cost}. 
For the cylindrical variant, we set
\begin{equation}
C_{\mathrm{Cyl}}(i,j)=1-\mathrm{IoU}_{\mathrm{cyl}}(i,j), \quad
A_{\mathrm{cyl}}(b)=\Delta\theta\Delta\phi .
\end{equation}
For the spherical variant, we use
\begin{equation}
C_{\mathrm{sph}}(i,j)=1-\mathrm{IoU}_{\mathrm{sph}}(i,j),
\end{equation}
\begin{equation}
A_{\mathrm{sph}}(b)=\Delta\theta(\sin\phi^{+}-\sin\phi^{-}),
\phi^{\pm}=\phi\pm\frac{\Delta\phi}{2}.
\end{equation}
The angle-free variant is obtained by setting \(\lambda_{\theta}=0\), while the raw-depth variant replaces the Kalman-predicted depth with the last observed trajectory depth.

The results show that spherical modeling is fully compatible with our framework but does not consistently outperform the proposed horizontal-periodic pixel-vertical cost in the considered setting. 
With the angular consistency term, the spherical variant achieves competitive IDF1, whereas the horizontal-periodic pixel-vertical representation obtains the best HOTA, AssA, and MOTA. 
This indicates that, for ERP-based ground robotic MOT, horizontal seam continuity is the dominant association factor, while vertical geometry can be sufficiently characterized by image-space boxes and depth cues. 
Therefore, we use the horizontal-periodic pixel-vertical cost as the default choice for online efficiency and task alignment, while retaining spherical overlap as a compatible option for scenarios with stronger vertical viewpoint changes or explicit 3D motion requirements.

\subsection{Ablation Analysis of Depth-Aware Spatio-Temporal Modeling}
Table~\ref{tab:ablation_depth_module} analyzes the contribution and coupling effects of depth-related designs under fixed geometry settings. Rows \ding{172}--\ding{174} show that instance-level depth provides complementary spatial cues for trajectory discrimination without changing the standard TBD pipeline. 
Further enabling Depth-Temporal Trajectory Modeling (DTM) in row \ding{174} achieves the best overall performance, demonstrating that trajectory-level depth refinement improves the temporal reliability of depth cues.

Rows \ding{175}--\ding{177} validate the importance of temporal consistency in depth-aware association. Compared with frame-wise association in row \ding{175}, both DTM in row \ding{176} and T-Mixer in row \ding{177} improve association-related metrics, showing that long-term trajectory consistency and batch-level temporal query interaction help stabilize depth cues over time.

Rows \ding{178}--\ding{180}, together with the full model in row \ding{174}, reveal the coupling between spatial and temporal consistency modeling.
SGA improves the frame-wise baseline by injecting spherical geometric priors, and combining SGA with DTM or T-Mixer outperforms temporal modeling alone in association performance.

The full configuration \ding{174} achieves the best HOTA, IDF1, AssA, and MOTA, suggesting that spatial priors and temporal consistency jointly improve the reliability of depth cues for panoramic MOT. 
The occlusion cases in Fig.~\ref{fig:vis_obstruction} further confirm that DTM helps separate closely overlapping targets with temporally smoothed depth cues, leading to more stable identity association under heavy occlusion.

\input{tables/params_analysis}
\input{figures/params_sensitive}
\input{tables/efficiency_anaysis}
\subsection{Efficiency Analysis}
We further evaluate the efficiency of \(\mathrm{CylindTrack}\) in terms of model complexity and inference speed. As shown in Table~\ref{tab:supp-end-to-end-complexity} and Table~\ref{tab:supp-sga-complexity}, \(\mathrm{CylindTrack}_{\mathrm{Det}}\) slightly increases the parameter count from $67.481M$ to $68.469M$, corresponding to a $1.46\%$ increase. Meanwhile, GMACs decrease from $123.888$ to $108.309$ and GFLOPs from $247.775$ to $216.618$, yielding reductions of $12.58\%$ and $12.57\%$, respectively.

The overall computational reduction mainly comes from the redesigned depth feature projection. Instead of projecting dense memory features, we perform temporal projection in the compact query space, allowing attention and feature transformation to operate on object-level representations and thereby avoiding redundant dense computation.
\input{figures/boundary_conditions}
SGA is an additional attention module introduced to explicitly model omnidirectional periodic geometric context with limited computational overhead. To this end, we replace direct full-context geometry attention with width--height separated context aggregation, compressing the geometry context from \(\sum_m H_mW_m\) tokens to \(\sum_m(H_m+W_m)\). This yields context compression factors of $36.46\times$ on QuadTrack and $39.89\times$ on JRDB. Even with two sequential directional attention calls, the resulting attention-core cost is $18.23\times$ and $19.95\times$ smaller than a conventional full-context implementation, respectively. In addition, the two directional branches use the same query projections while sharing the key/value and output projections, which further limits the parameter overhead. Consequently, the proposed implementation enables SGA to retain full omnidirectional periodic awareness while introducing only a small additional computational and parameter cost to the baseline.

The practical FPS results further support this design. While improving the main tracking metrics, including HOTA, IDF1, and AssA, \(\mathrm{CylindTrack}_{\mathrm{Det}}\) maintains comparable or even higher FPS than the baseline. In addition, the association module transforms the original association state into a topology-consistent state space without introducing additional learned modules or noticeable computational overhead. Therefore, the downstream association process preserves the decoupled and lightweight nature of existing trackers while benefiting from topology-aware matching cues.

\subsection{Parameter Analyses}
As shown in Fig.~\ref{fig:params}, the proposed configuration is consistently stable across different settings. 
We analyze key hyperparameters related to depth modeling, cylindrical geometric consistency, and association. 
$\lambda_z$, in Eq.~\ref{eq:depth_assoc_cost}, controls the relative contribution of depth consistency in the association cost.

For association, $\tau_m$ defines the matching threshold for accepting trajectory-detection pairs, and $\tau_d$ acts as a depth-gating threshold to reject matches with large predicted-observed depth discrepancies. 
In addition, $\lambda_\theta$, in Eq.~\ref{eq:assoc_cost_params}, weights the cylindrical geometric consistency cost as an auxiliary association cue. 
For CylindTrack, $\tau_h$ and $\tau_l$ denote the high and low detection confidence thresholds used in the first-stage and second-stage association, respectively, and the new-track initialization threshold is set as $\tau_{\mathrm{new}}=\tau_h+0.10$.
The best or near-best performance is obtained around the default setting, \textit{i.e.}, $\lambda_z=0.2$, matching threshold $0.8$, and $\lambda_\theta=0.1$. 
Extremely weak or overly strong association constraints lead to clear degradation, while moderate perturbations around the default configuration preserve competitive performance. This indicates that our method is not tuned to a single detector or an isolated hyperparameter choice.

\subsection{Failure Case Analyses}

The boundary problem discussed in this section involves both coordinate discontinuity and fragmented object observations. TCMM addresses the former by lifting horizontal motion into an unwrapped angular state and incorporating periodic overlap and angular consistency into association. However, continuity in the tracking state does not necessarily imply continuity in the detector's observations: a single physical target may still occupy two disconnected regions of the ERP image, receive incomplete annotations, or produce poorly localized detections near the seam.
Fig.~\ref{fig:boundary_conditions} examines these complementary boundary conditions. Panel (a) documents boundary-crossing events and their annotation coverage, highlighting a recurring tracking condition that warrants explicit modeling and dedicated evaluation. These observations motivate BCIC as a local measure of identity consistency across panoramic boundaries, while panels (b)-(d) illustrate the annotation and detection difficulties surrounding such transitions. Panel (b) shows a visible target whose annotation disappears temporarily near the seam. Panels (c) and (d) show examples from CylindTrack in which detection-overlap degradation and interrupted matching precede identity reassignment. These observations complement BCIC by exposing the annotation and detection conditions surrounding identity discontinuities.

CylindTrack takes a step toward a more general panoramic representation and tracking paradigm by aligning depth-aware instance representations and trajectory dynamics with panoramic geometry. Within SSTC, SGA injects spherical geometric priors into instance-level depth refinement, while the query-based Temporal Mixer improves the local temporal coherence of depth queries. DTM then propagates depth as a trajectory-level state, and TCMM maintains horizontal continuity during motion prediction and association. Together, these designs mitigate geometric and temporal inconsistencies, particularly by enabling the tracker to model horizontal motion in a continuous angular state space and preserve trajectory and identity continuity across the panoramic seam through topology-consistent prediction and association. However, this tracker-level periodic modeling does not explicitly unify seam-separated visual fragments into a coherent object representation at the detection stage.

Further progress therefore requires extending panoramic consistency to the detection and supervision stages. Cross-seam visual feature aggregation and unified instance prediction should be considered alongside annotation processing that ensures continuity of object identities and spatial extents across the left-right boundary. Such designs would complement the proposed depth refinement and cylindrical association framework by reducing fragmentation at its observation source. We leave this joint treatment of panoramic detection and annotation consistency to future work.

%% file: figures/lift_to_3d.tex
\begin{figure}[t]
    \centering
    \includegraphics[
        width=0.48\textwidth,
        trim=105pt 0pt 60pt 0pt,
        clip
    ]{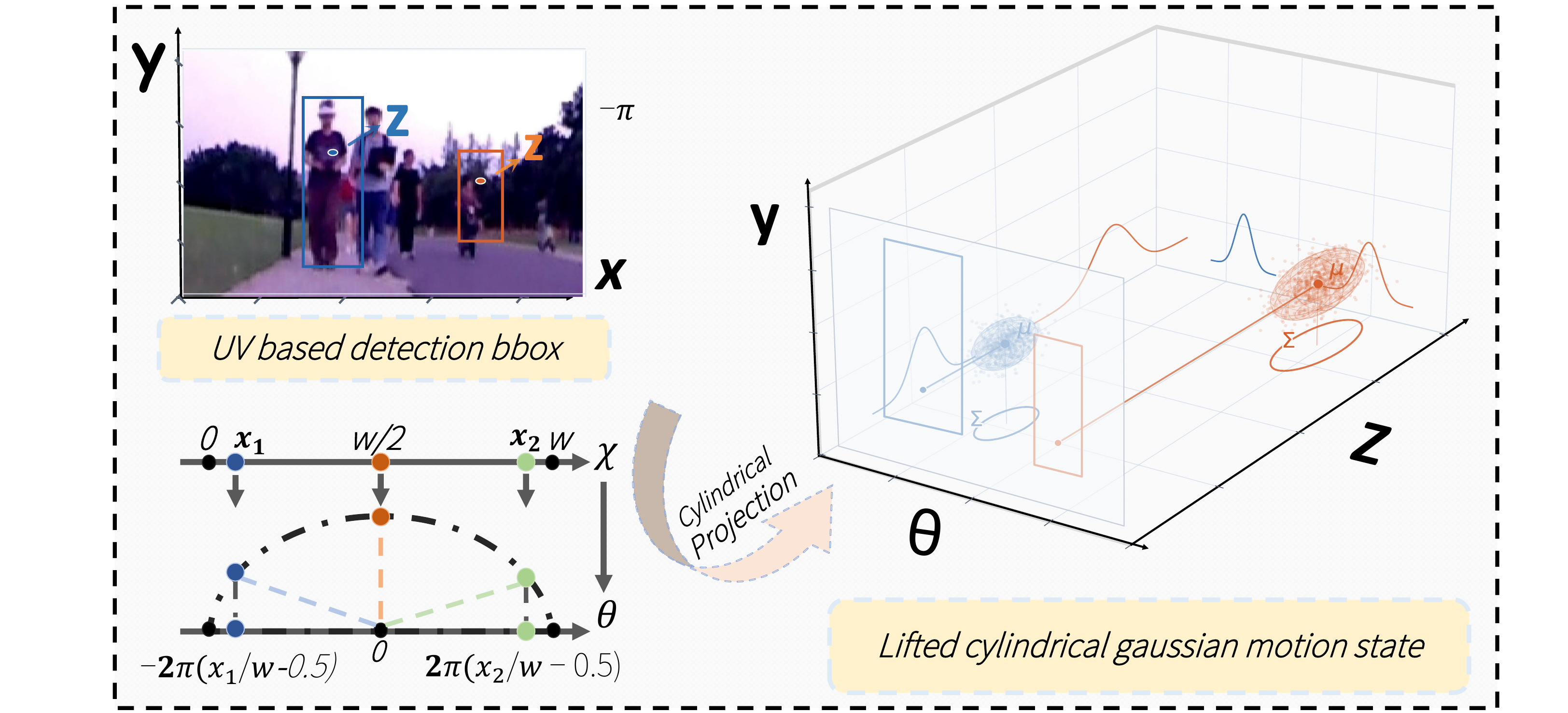}
    \caption{Visualization of the proposed cylindrical lifting. Equirectangular detections are mapped from horizontal image coordinates to periodic longitude angles, and their depths are converted into radial distances. In this way, ERP detections are lifted from image-space boxes to a depth-aware cylindrical state for topology-consistent trajectory association.
    }
    \label{fig:lifting}
\end{figure}

%% file: tables/main_JRDB_table.tex
\begin{table*}[t]
\centering
\small
\setlength{\tabcolsep}{4.2pt}
\renewcommand{\arraystretch}{0.8}

\newcommand{\jrdbposgain}[1]{\textcolor{green!50!black}{\scriptsize\,\ensuremath{(+#1)}}}
\newcommand{\jrdbneggain}[1]{\textcolor{gray}{\scriptsize\,\ensuremath{(#1)}}}

\resizebox{\textwidth}{!}{
\begin{tabular}{l|c|rrrrrr}
\toprule[1.2pt]
\textbf{Tracker} & \textbf{Detector}
& \textbf{HOTA$\uparrow$}
& \textbf{BCIC$\uparrow$}
& \textbf{IDF1$\uparrow$}
& \textbf{MOTA$\uparrow$}
& \textbf{AssA$\uparrow$}
& \textbf{FPS$\uparrow$} \\
\midrule

\multirow{2}{*}{SORT~\cite{bewley2016sort}}
& DepTR-MOT
& 22.992 & 26.803 & 22.941 & 32.327 & 18.125 & 21.732 \\
& \(\mathrm{CylindTrack}_{\mathrm{Det}}\)
& 23.127\jrdbposgain{0.135}
& 25.126\jrdbneggain{-1.677}
& 23.032\jrdbposgain{0.091}
& 30.750\jrdbneggain{-1.577}
& 19.492\jrdbposgain{1.367}
& 21.469\jrdbneggain{-0.263} \\
\midrule

\multirow{2}{*}{DeepSORT~\cite{wojke2017deepsort}}
& DepTR-MOT
& 26.725 & 27.982 & 28.537 & \textbf{32.869} & 23.039 & 18.938 \\
& \(\mathrm{CylindTrack}_{\mathrm{Det}}\)
& 27.459\jrdbposgain{0.734}
& 28.835\jrdbposgain{0.853}
& 29.243\jrdbposgain{0.706}
& 32.472\jrdbneggain{-0.397}
& 25.484\jrdbposgain{2.445}
& 19.243\jrdbposgain{0.304} \\
\midrule

\multirow{2}{*}{Bot-SORT~\cite{aharon2022botsortrobustassociationsmultipedestrian}}
& DepTR-MOT
& 21.754 & 27.524 & 20.850 & -8.040 & 17.316 & 20.554 \\
& \(\mathrm{CylindTrack}_{\mathrm{Det}}\)
& 25.331\jrdbposgain{3.577}
& 29.613\jrdbposgain{2.089}
& 25.460\jrdbposgain{4.610}
& 19.988\jrdbposgain{28.028}
& 20.969\jrdbposgain{3.653}
& 20.962\jrdbposgain{0.408} \\
\midrule

\multirow{2}{*}{OC-SORT~\cite{cao2023observation}}
& DepTR-MOT
& 20.613 & 26.450 & 21.356 & 26.248 & 17.347 & \textbf{24.909} \\
& \(\mathrm{CylindTrack}_{\mathrm{Det}}\)
& 20.988\jrdbposgain{0.375}
& 24.668\jrdbneggain{-1.782}
& 21.654\jrdbposgain{0.298}
& 25.120\jrdbneggain{-1.128}
& 19.159\jrdbposgain{1.812}
& 23.328\jrdbneggain{-1.581} \\
\midrule

\multirow{2}{*}{HybridSORT~\cite{yang2024hybrid}}
& DepTR-MOT
& 18.091 & 28.330 & 19.646 & 18.024 & 19.748 & 16.743 \\
& \(\mathrm{CylindTrack}_{\mathrm{Det}}\)
& 18.048\jrdbneggain{-0.043}
& 27.329\jrdbneggain{-1.001}
& 19.172\jrdbneggain{-0.474}
& 17.243\jrdbneggain{-0.781}
& 20.969\jrdbposgain{1.221}
& 15.674\jrdbneggain{-1.069} \\
\midrule

\multirow{2}{*}{DiffMOT~\cite{lv2024diffmot}}
& DepTR-MOT
& 20.142 & 27.169 & 19.437 & -15.617 & 15.395 & 20.439 \\
& \(\mathrm{CylindTrack}_{\mathrm{Det}}\)
& 23.716\jrdbposgain{3.574}
& 27.404\jrdbposgain{0.235}
& 23.746\jrdbposgain{4.309}
& 15.511\jrdbposgain{31.128}
& 18.912\jrdbposgain{3.517}
& 20.895\jrdbposgain{0.456} \\
\midrule

\multirow{2}{*}{ByteTrack~\cite{zhang2022bytetrack}}
& DepTR-MOT 
& 28.392 & 21.852 & 31.025 & 31.631 & 26.399 & 19.058 \\
& \(\mathrm{CylindTrack}_{\mathrm{Det}}\)
& 28.648\jrdbposgain{0.256}
& 28.177\jrdbposgain{6.325}
& 31.552\jrdbposgain{0.527}
& 31.702\jrdbposgain{0.071}
& 28.303\jrdbposgain{1.904}
& 17.401\jrdbneggain{-1.657} \\
\midrule[\heavyrulewidth]

\multicolumn{2}{c|}{\(\mathrm{OmniTrack}_{\mathrm{TBD}}\)~\cite{luo2025omnitrack}}
& 26.759
& 21.383
& 29.705
& 27.087
& 29.519
& 20.384 \\
\midrule

\multicolumn{2}{c|}{\(\mathrm{CylindTrack}_{\mathrm{TBD}}\)}
& \textbf{31.117}
& \textbf{35.365}
& \textbf{34.331}
& 31.219
& \textbf{31.347}
& 21.339 \\
\bottomrule[1.2pt]
\end{tabular}
}
\vskip-1ex
\caption{Depth-enabled tracking results using DepTR-MOT~\cite{deng2025deptrmot} and \(\mathrm{CylindTrack}_{\mathrm{Det}}\) as detectors on the JRDB test set~\cite{martin2021jrdb}.
The colored values in parentheses denote the absolute performance gain of \(\mathrm{CylindTrack}_{\mathrm{Det}}\) over DepTR-MOT. BCIC denotes Boundary Crossing Identity Consistency, a boundary-aware metric defined in Section~\ref{sec:Metrics}.
}
\label{tab:depth-detector-comparison}
\end{table*}

%% file: tables/ablation.tex
\begin{table}[!t]
\centering
\caption{Ablation study of main designs on CylindTrack.}
\label{tab:ablation}

\vspace{-2mm}
\setlength{\tabcolsep}{2.4pt}
\renewcommand{\arraystretch}{0.90}
\tiny

\resizebox{\columnwidth}{!}{
\begin{tabular}{c ccc ccccc}
\toprule
\multirow{2}{*}{\textbf{ID}} 
& \multicolumn{3}{c}{\textbf{Comp.}} 
& \multicolumn{5}{c}{\textbf{Metrics}} \\
\cmidrule(lr){2-4} \cmidrule(lr){5-9}
& A & B & C
& HOTA$\uparrow$ & IDF1$\uparrow$ & AssA$\uparrow$ & DetA$\uparrow$ & MOTA$\uparrow$ \\
\midrule

\eid{1}
& \xmark & \xmark & \xmark
& 31.026 & 34.109 & 29.786 & 33.497 & 14.387 \\

\eid{2}
& \cmark & \xmark & \xmark
& 31.387 & 35.025 & 30.322 & 33.587 & 14.843 \\

\eid{3}
& \xmark & \cmark & \xmark
& 30.009 & 34.641 & 27.702 & 33.320 & 20.046 \\

\eid{4}
& \xmark & \xmark & \cmark
& 31.464 & 36.546 & 30.706 & 33.408 & 13.239\\

\arrayrulecolor{gray!45}\midrule\arrayrulecolor{black}

\eid{5}
& \cmark & \xmark & \cmark
& 31.930 & 37.520 & 31.560 & 33.420 & 13.787 \\

\eid{6}
& \xmark & \cmark & \cmark
& 33.000 & 39.645 & 33.206 & \best{33.608} & 20.253 \\

\eid{7}
& \cmark & \cmark & \xmark
& 30.817 & 35.498 & 29.255 & 33.341 & 20.334 \\

\arrayrulecolor{gray!45}\midrule\arrayrulecolor{black}

\eid{8}
& \cmark & \cmark & \cmark
& \best{33.674} & \best{40.446} & \best{34.665} & 33.341 & \best{20.583} \\
\bottomrule
\end{tabular}
}

\vspace{1.5mm}
\begin{minipage}{0.98\columnwidth}

\small
\textbf{Components.} 
A: Depth-Temporal Trajectory Modeling (DTM);
B: Spherical Spatio-Temporal Consistency (SSTC);
C: Topology-Aware Cylindrical Motion Model (TCMM).
\end{minipage}

\end{table}

%% file: tables/ablation_5.tex
\begin{table}[t]
\centering
\caption{Ablation study of different tracker components.}
\label{tab:cyl_sph_ablation}
\resizebox{\columnwidth}{!}{
\begin{tabular}{c c c c c c c c c}
\toprule
ID & $\mathcal{C}$  & $\mathcal{T}$  & $\mathcal{G}$  & $\mathcal{A}$  & HOTA$\uparrow$ & IDF1$\uparrow$ & AssA$\uparrow$ & MOTA$\uparrow$ \\
\midrule
\ding{172} &     &     & Img &     & 30.817 & 35.498 & 29.255 & 20.334 \\
\ding{173} & \cmark &     & Img &     & 30.893 & 35.701 & 29.390 & 20.268 \\
\ding{174} &     & \cmark & Cyl &     & 30.818 & 35.489 & 29.271 & 20.331 \\
\ding{175} & \cmark & \cmark & Cyl &     & 31.616 & 37.026 & 30.794 & 20.154 \\
\ding{176} &     &     & Img & \cmark & 32.700 & 38.381 & 32.810 & 20.232 \\
\ding{177} & \cmark & \cmark & Cyl & \cmark & 33.659 & 40.412 & 34.622 & 20.557 \\
\ding{178} & \cmark & \cmark & Sph &     & 31.536 & 36.756 & 30.532 & 20.531 \\
\ding{179} & \cmark & \cmark & Sph & \cmark & 33.480 & \textbf{40.560} & 34.312 & 20.245 \\
\ding{180} & \cmark & \cmark & HPV & \cmark & \textbf{33.674} & 40.446 & \textbf{34.665} & \textbf{20.583} \\
\bottomrule
\end{tabular}
}
\begin{flushleft}
\footnotesize
\textbf{Components.}
$\mathcal{C}$, $\mathcal{T}$,$\mathcal{G}$, and $\mathcal{A}$ denote cylindrical motion Kalman filter, topology-aware IoU, geometry type, and periodic angle cost, respectively. 
Img, Cyl, Sph, and HPV denote the overlap terms computed using image-plane IoU, cylindrical IoU, spherical-area IoU, and horizontal-periodic pixel-vertical IoU, respectively. All rows use the same DTM setting. Best results are highlighted in bold.
\end{flushleft}
\end{table}

%% file: tables/ablation_depth.tex
\begin{table}[!t]
\centering
\caption{Depth-related ablations on CylindTrack.}
\resizebox{\columnwidth}{!}{
\begin{tabular}{@{}c c c c c c c c c@{}}
\toprule
ID & Depth & DTM & SGA & T-Mixer
& HOTA$\uparrow$ & IDF1$\uparrow$ & AssA$\uparrow$ & MOTA$\uparrow$ \\
\midrule

\ding{172} 
&     &     & \cmark & \cmark   
& 29.999 & 34.191 & 27.722 & 19.682 \\

\ding{173} 
& \cmark &     & \cmark & \cmark  
& 33.000 & 39.645 & 33.206 & 20.253 \\

\ding{174} 
& \cmark & \cmark & \cmark & \cmark 
& \textbf{33.674} & \textbf{40.446} & \textbf{34.665} & \textbf{20.583} \\

\midrule

\ding{175}
& \cmark &  &  &  
& 31.026 & 34.109 & 29.786 & 14.387 \\

\ding{176} 
& \cmark & \cmark & & 
& 31.930 & 37.520 & 31.560 & 13.787 \\

\ding{177} 
& \cmark &  &  & \cmark
& 31.314 & 37.148 & 30.851 & 16.728 \\

\midrule

\ding{178} 
& \cmark &  & \cmark & 
& 32.347 & 38.537 & 32.017 & 17.575 \\

\ding{179} 
& \cmark & \cmark & \cmark &     
& 32.244 & 38.539 & 32.042 & 17.068 \\

\ding{180} 
& \cmark & \cmark &     & \cmark 
& 31.464 & 36.735 & 31.058 & 17.293 \\
\bottomrule
\end{tabular}
}
\begin{flushleft}
\footnotesize
\textbf{Components.}
``Depth'' denotes the use of instance-level depth cues in trajectory association, DTM denotes Depth-Temporal Trajectory Modeling, SGA injects spherical geometric priors, and Temporal Mixer enhances temporal query consistency.
\label{tab:ablation_depth_module}
\end{flushleft}
\end{table}

%% file: figures/visualization_boundary.tex
\begin{figure*}[t]
    \centering

    \scalebox{1}[0.85]{%
    \includegraphics[
        width=0.98\textwidth,
        trim={0 0 0 55mm}, 
        clip
    ]{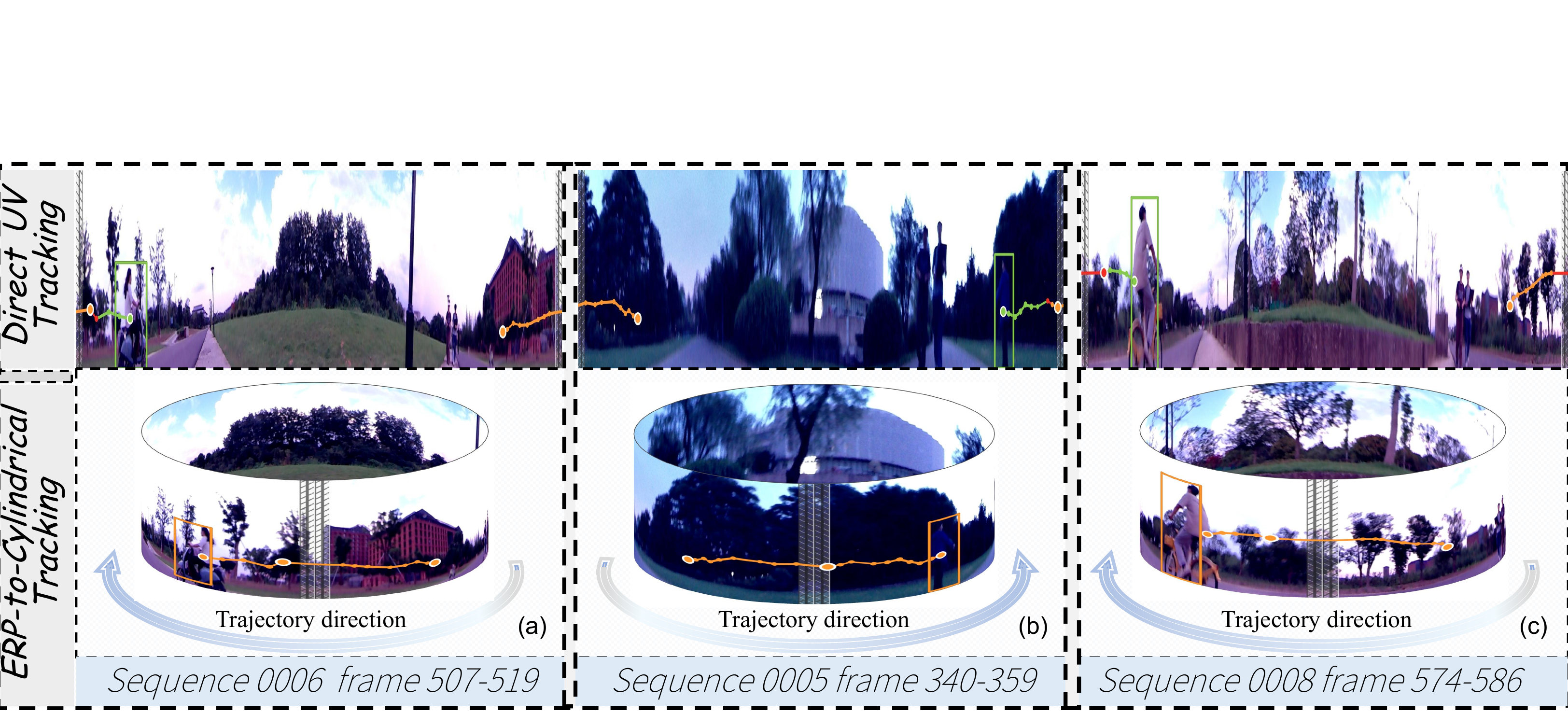}
    }

    \vspace{-0.8mm}

    \scalebox{1}[0.85]{%
    \includegraphics[
        width=0.98\textwidth,
        trim={0 0 0 58mm}, 
        clip
    ]{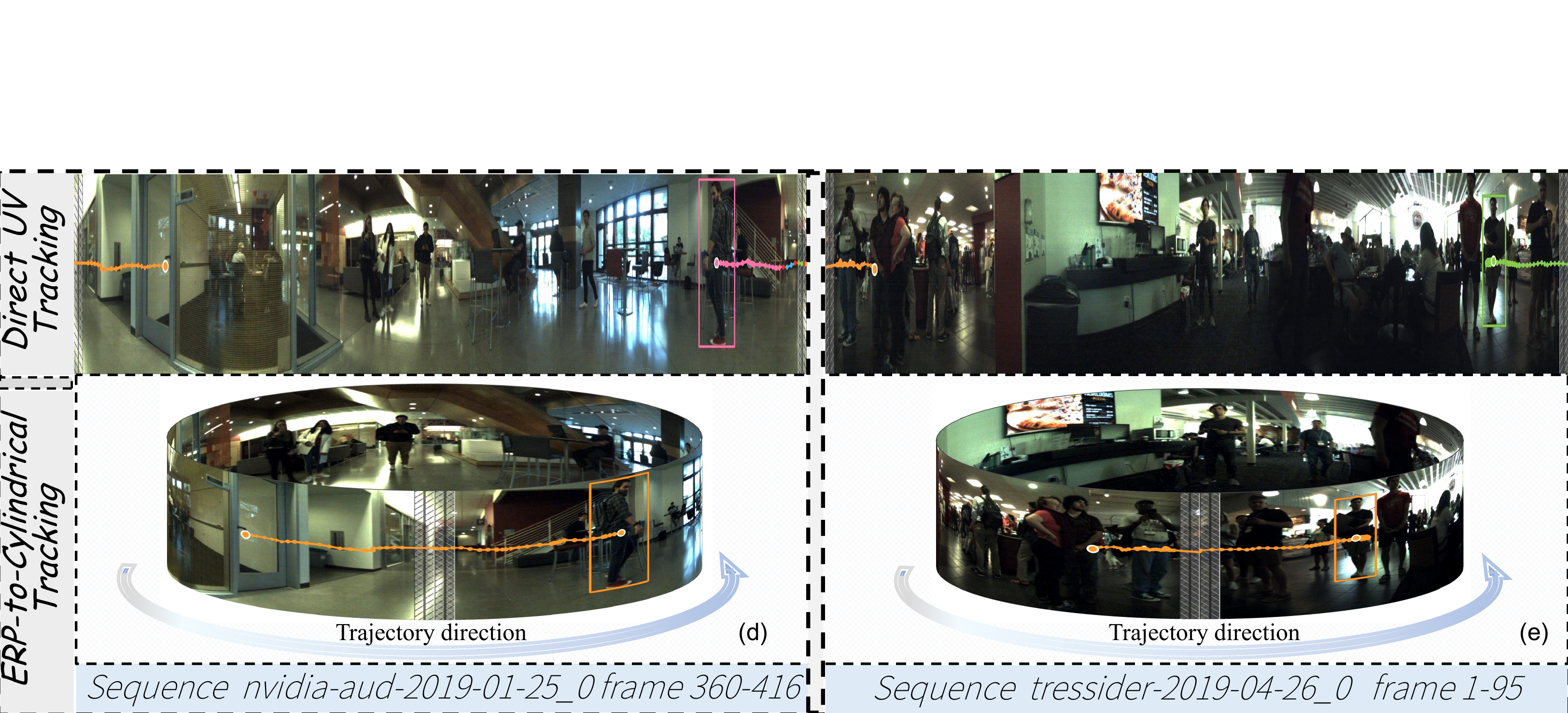}
    }
    \vskip-1ex
    \caption{Qualitative visualization of CylindTrack on the QuadTrack~\cite{luo2025omnitrack} (a/b/c) and JRDB~\cite{martin2021jrdb} (d/e) test sets.
    By leveraging the panoramic topology, CylindTrack maintains cross-boundary trajectory continuity when targets move across image boundaries, enabling consistent identity association under panoramic observations, effectively associating targets whose appearances and locations undergo periodic spatial transformations near image boundaries.}
    \label{fig:visualization_quadtrack_jrdb}
    \vskip-3ex
\end{figure*}

%% file: figures/visualization_quadtrack.tex
\begin{figure*}[!t]
    \centering
    \scalebox{1}[1]{%
    \includegraphics[
        width=0.98\textwidth,
        trim={0 0 0 90mm}, 
        clip
    ]{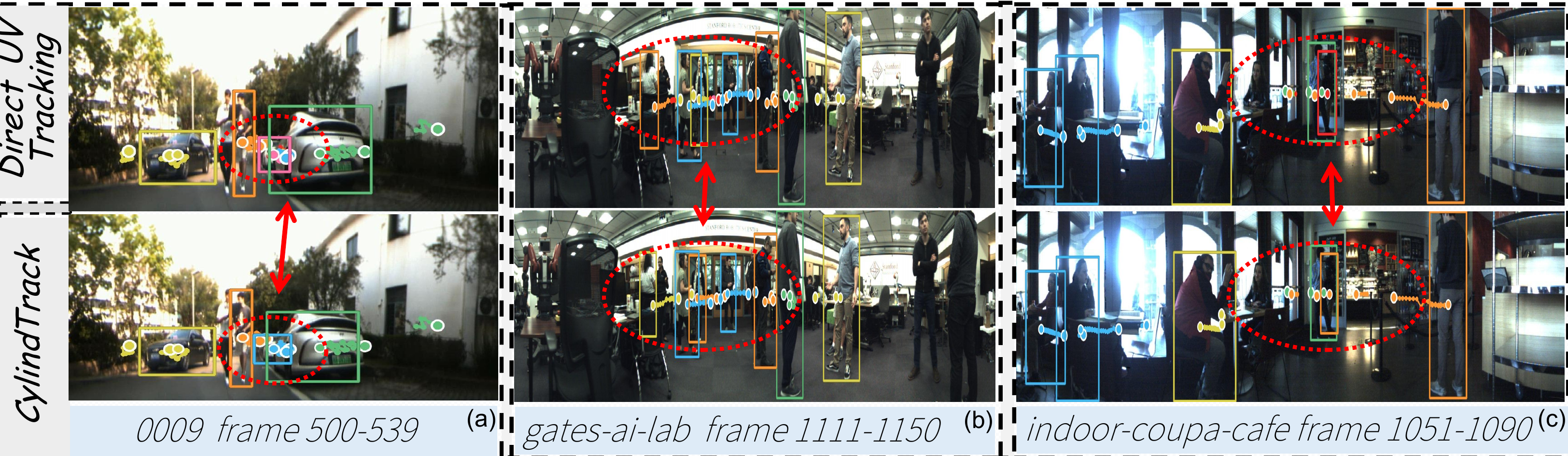}
    }
    \vskip-1ex
    \caption{Qualitative visualization of CylindTrack under heavy occlusion on the QuadTrack (a) and JRDB (b/c) test sets~\cite{luo2025omnitrack,martin2021jrdb}.
    Our Depth-Temporal Trajectory Modeling introduces discriminative 3D cues, improving identity separability for closely overlapping targets and supporting robust association in highly occluded scenes.
    }
    \label{fig:vis_obstruction}
    \vskip-4ex
\end{figure*}

%% file: tables/params_analysis.tex
\begin{table}[t]
\centering
\caption{Ablation of boundary window length $L$ on QuadTrack using BCIC.}

\label{tab:bcic_ablation_compact}
\vspace{-2mm}
\setlength{\tabcolsep}{3.2pt}
\renewcommand{\arraystretch}{0.85}
\footnotesize

\newcommand{\gain}[1]{\textcolor{green!50!black}{\scriptsize $(+#1)$}}

\resizebox{\columnwidth}{!}{
\begin{tabular}{lccccccc}
\toprule
\textbf{Detector}
& \textbf{$L=11$}
& \textbf{$L=21$}
& \textbf{$L=31$}
& \textbf{$L=61$}
& \textbf{$L=91$}
& \textbf{$L=121$}
& \textbf{$L=181$} \\
\midrule

DepTR-MOT
& 30.48
& 30.86
& 30.58
& 27.24
& 23.63
& 20.48
& 17.16 \\

\(\mathrm{CylindTrack_{TBD}}\)
& \textbf{51.07}
& \textbf{54.81}
& \textbf{54.96}
& \textbf{50.24}
& \textbf{43.70}
& \textbf{38.11}
& \textbf{30.41} \\

\midrule
\textbf{Gain}
& \gain{20.59}
& \gain{23.96}
& \gain{24.38}
& \gain{23.00}
& \gain{20.06}
& \gain{17.63}
& \gain{13.25} \\

\bottomrule
\end{tabular}
}
\begin{flushleft}
\footnotesize
\textbf{Components.}
The experiments report the results of \(\mathrm{CylindTrack}_{\mathrm{TBD}}\) and DepTR-MOT, where \(L=2\Delta+1\) denotes the boundary window length.
\end{flushleft}
\vspace{-2mm}
\end{table}

%% file: figures/params_sensitive.tex
\begin{figure}[t]
\centering

\includegraphics[
    width=0.48\textwidth,
    trim=0pt 0pt 0pt 0pt,
    clip
]{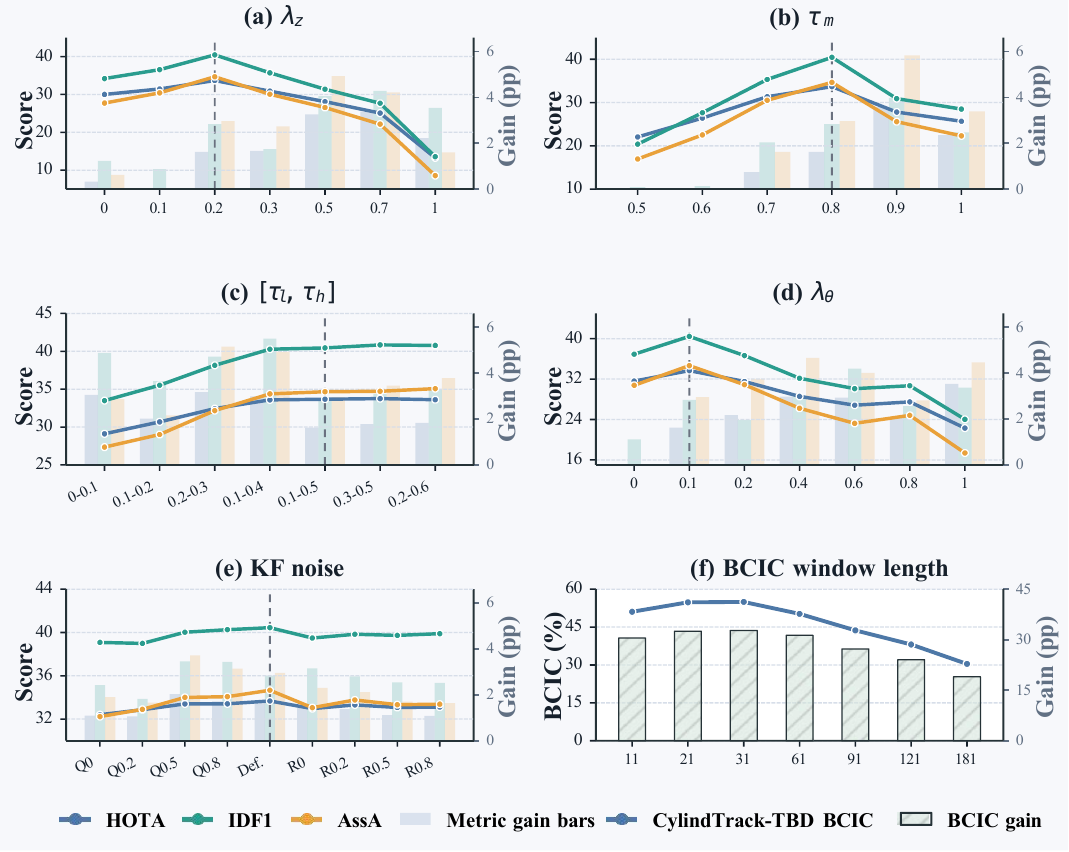}
\caption{Sensitivity analysis of \(\mathrm{CylindTrack}_{\mathrm{TBD}}\) association hyperparameters and BCIC boundary-window ablation on the QuadTrack test set. 
Panels (a)--(e) show the sensitivity to \(\lambda_z\), \(\tau_m\), \([\tau_l,\tau_h]\), \(\lambda_\theta\), and the Kalman-filter noise scale, respectively. 
Solid curves report HOTA, IDF1, and AssA, shaded bars indicate absolute gains over DepTR-MOT, and dashed lines mark the selected/default settings. 
Panel (f) reports the BCIC score and gain under different boundary-window lengths \(L=2\Delta+1\).}
\label{fig:params}
\end{figure}

%% file: tables/efficiency_anaysis.tex
\begin{table}[t]
\centering
\caption{End-to-end detector complexity.}
\label{tab:supp-end-to-end-complexity}

\small
\setlength{\tabcolsep}{2.5pt}
\renewcommand{\arraystretch}{1.0}

\begin{tabular}{@{}lccc@{}}
\toprule
\textbf{Model} & \textbf{Params (M)} & \textbf{GMACs} & \textbf{GFLOPs} \\
\midrule
Original detector & 67.481 & 123.888 & 247.775 \\
$\mathrm{CylindTrack}_{\mathrm{Det}}$ & 68.469 & 108.309 & 216.618 \\
\midrule
Absolute change & $+0.988$ & $-15.579$ & $-31.157$ \\
Relative change & $+1.46\%$ & $-12.58\%$ & $-12.57\%$ \\
\bottomrule
\end{tabular}

\end{table}

\begin{table*}[t]
\centering
\caption{Theoretical cost of one conventional full-context attention call and
the implemented width--height separated SGA with two sequential calls. GFLOPs
count only $\mathbf{Q}\mathbf{K}^{\top}$ and
$\operatorname{Softmax}(\cdot)\mathbf{V}$. Score memory is the inference-time
peak for one explicit FP32 map per attention head; K/V memory is the shared FP32 cache
over all channels. FP16 values are one half. ``Full/Impl.'' reports the two
alternatives.}
\vspace{-2mm}
\label{tab:supp-sga-complexity}
\scriptsize
\setlength{\tabcolsep}{4.5pt}
\renewcommand{\arraystretch}{1}
\resizebox{\textwidth}{!}{%
\begin{tabular}{lcccccc}
\toprule
\textbf{Dataset} & $\boldsymbol{N_s}$ & $\boldsymbol{N_{\mathrm{sep}}}$ &
\textbf{GFLOPs Full/Impl.} & \textbf{Saved GFLOPs} &
\textbf{Peak Score Mem. (MiB/h), F/I} & \textbf{K/V MiB Full/Impl.} \\
\midrule
QuadTrack & 20,160 & 553 & 6.193 / 0.340 & 5.853 & 23.07 / 0.63 & 39.38 / 1.08 \\
JRDB      & 37,020 & 928 & 11.373 / 0.570 & 10.802 & 42.37 / 1.06 & 72.30 / 1.81 \\
\bottomrule
\end{tabular}}
\vspace{-8pt}
\end{table*}

%% file: figures/boundary_conditions.tex
\begin{figure}[t]
    \centering
    \includegraphics[
        width=\columnwidth,
        trim=10 20 10 20,
        clip
    ]
    {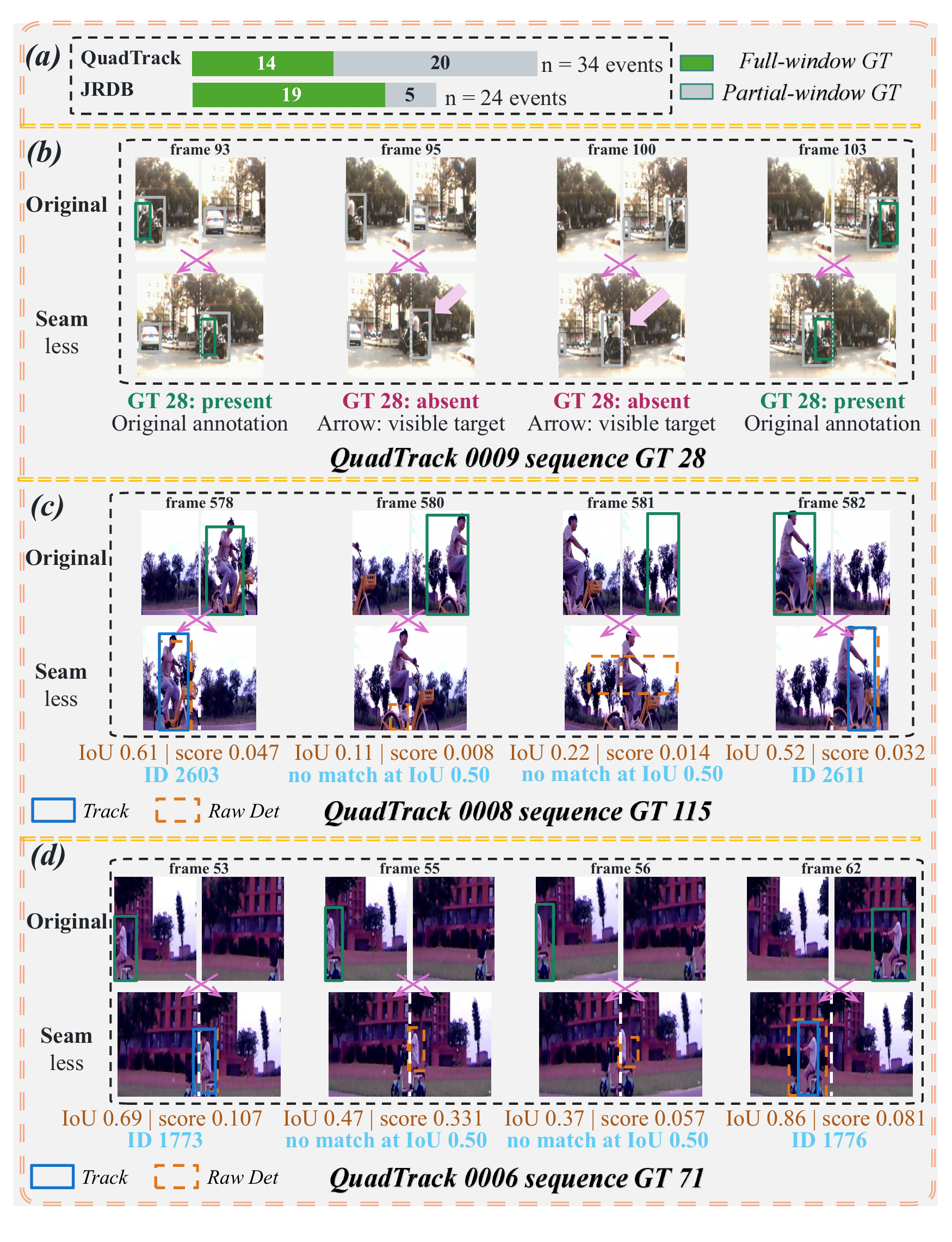}
    \vspace{-3ex}
    \caption{Boundary conditions visualization in panoramic tracking. (a) GT coverage within 61-frame visualization windows: 14/20 boundary-crossing events on QuadTrack (\(n=34\)) and 19/5 on JRDB (\(n=24\)); these windows provide descriptive evidence context for BCIC. (b) A visible target with interrupted annotation in QuadTrack 0009, GT 28: the target remains visible in frames 94--102 despite a gap in the annotations. (c) A target split between the left and right image edges
in QuadTrack 0008, GT 115: the target splits across the image edges, detection overlap decreases, and the matched identity changes from 2603 to 2611. (d) A second example in sequence 0006 (GT 71), showing a matched identity of 1773 at frame 53, subthreshold raw detection overlap and no matched tracking output at frames 55-56, and identity 1776 at frame 62. In (c)-(d), orange dashed outlines show only the raw detection candidate with the highest circular IoU with the annotated target, with its recorded confidence printed below; blue solid outlines show matched tracking outputs with circular $\mathrm{IoU}\geq 0.50$. The Original row preserves the left and right image boundaries of the Original images, and the Seamless row stitches the left and right parts together to display continuous image content and trajectories.}
    \label{fig:boundary_conditions}
\end{figure}

%% file: body/conclusion.tex
\section{Conclusion}
CylindTrack combines periodic angular motion modeling with temporally filtered relative-depth cues for panoramic multi-object tracking. The results demonstrate improved identity consistency and trajectory continuity across the panorama seam under the evaluated conditions, supporting persistent multi-object tracking for mobile robot perception. It is worth noting that the DTM state is designed to model temporally smoothed relative depth in the panoramic image stream, rather than absolute metric depth in a world coordinate system. This lightweight formulation is well aligned with online tracking-by-detection, as it avoids reliance on external pose estimation or full metric reconstruction. Future work will extend CylindTrack toward camera-motion-aware depth filtering, more general 360\textdegree{} tracking scenarios with diverse sensors and object categories, and more tightly coupled architectures that jointly learn seam-consistent detection and camera-motion-aware depth perception, under panoramic geometry.

%% file: main.bbl
\begin{thebibliography}{10}
\providecommand{\url}[1]{#1}
\csname url@samestyle\endcsname
\providecommand{\newblock}{\relax}
\providecommand{\bibinfo}[2]{#2}
\providecommand{\BIBentrySTDinterwordspacing}{\spaceskip=0pt\relax}
\providecommand{\BIBentryALTinterwordstretchfactor}{4}
\providecommand{\BIBentryALTinterwordspacing}{\spaceskip=\fontdimen2\font plus
\BIBentryALTinterwordstretchfactor\fontdimen3\font minus \fontdimen4\font\relax}
\providecommand{\BIBforeignlanguage}[2]{{%
\expandafter\ifx\csname l@#1\endcsname\relax
\typeout{** WARNING: IEEEtran.bst: No hyphenation pattern has been}%
\typeout{** loaded for the language `#1'. Using the pattern for}%
\typeout{** the default language instead.}%
\else
\language=\csname l@#1\endcsname
\fi
#2}}
\providecommand{\BIBdecl}{\relax}
\BIBdecl

\bibitem{martin2021jrdb}
R.~Mart{\'{\i}}n{-}Mart{\'{\i}}n \emph{et~al.}, ``{JRDB:} {A} dataset and benchmark of egocentric robot visual perception of humans in built environments,'' \emph{IEEE Transactions on Pattern Analysis and Machine Intelligence}, vol.~45, no.~6, pp. 6748--6765, 2023.

\bibitem{luo2025omnitrack}
K.~Luo \emph{et~al.}, ``Omnidirectional multi-object tracking,'' in \emph{Proc. CVPR}, 2025, pp. 21\,959--21\,969.

\bibitem{bewley2016sort}
A.~Bewley, Z.~Ge, L.~Ott, F.~Ramos, and B.~Upcroft, ``Simple online and realtime tracking,'' in \emph{Proc. ICIP}, 2016, pp. 3464--3468.

\bibitem{wojke2017deepsort}
N.~Wojke, A.~Bewley, and D.~Paulus, ``Simple online and realtime tracking with a deep association metric,'' in \emph{Proc. ICIP}, 2017, pp. 3645--3649.

\bibitem{yang2024hybrid}
M.~Yang \emph{et~al.}, ``{Hybrid-SORT:} {Weak} cues matter for online multi-object tracking,'' in \emph{Proc. AAAI}, 2024, pp. 6504--6512.

\bibitem{lv2024diffmot}
W.~Lv, Y.~Huang, N.~Zhang, R.-S. Lin, M.~Han, and D.~Zeng, ``{DiffMOT:} {A} real-time diffusion-based multiple object tracker with non-linear prediction,'' in \emph{Proc. CVPR}, 2024, pp. 19\,321--19\,330.

\bibitem{aharon2022botsortrobustassociationsmultipedestrian}
N.~Aharon, R.~Orfaig, and B.-Z. Bobrovsky, ``{BoT-SORT:} {Robust} associations multi-pedestrian tracking,'' \emph{arXiv:2206.14651}, 2022.

\bibitem{shenoi2020jrmotrealtime3dmultiobject}
A.~Shenoi \emph{et~al.}, ``{JRMOT:} {A} real-time {3D} multi-object tracker and a new large-scale dataset,'' in \emph{Proc. IROS}, 2020, pp. 10\,335--10\,342.

\bibitem{wang2024sttracknet}
S.~Wang, Y.~Sun, Z.~Wang, and M.~Liu, ``{ST-TrackNet}: {A} multiple-object tracking network using spatio-temporal information,'' \emph{IEEE Transactions on Automation Science and Engineering}, vol.~21, no.~1, pp. 284--295, 2024.

\bibitem{yan2026chtracker}
C.~Yan, Y.~Wang, Y.~Qing, W.~Zhang, and X.~Xu, ``{CHTracker}: {Confidence-guided} hierarchical association paradigm for multi-object tracking,'' \emph{IEEE Transactions on Automation Science and Engineering}, vol.~23, pp. 1816--1829, 2026.

\bibitem{muzaddid2024ntrack}
M.~A. Al~Muzaddid and W.~J. Beksi, ``{NTrack}: {A} multiple-object tracker and dataset for infield cotton boll counting,'' \emph{IEEE Transactions on Automation Science and Engineering}, vol.~21, no.~4, pp. 7452--7464, 2024.

\bibitem{deng2025deptrmot}
B.~Deng, L.~Huang, K.~Luo, F.~Teng, and K.~Yang, ``{DepTR-MOT:} {Unveiling} the potential of depth-informed trajectory refinement for multi-object tracking,'' in \emph{Proc. ROBIO}, 2026.

\bibitem{liu2025sparsetrack}
Z.~Liu, X.~Wang, C.~Wang, W.~Liu, and X.~Bai, ``{SparseTrack:} {Multi-object} tracking by performing scene decomposition based on pseudo-depth,'' \emph{IEEE Transactions on Circuits and Systems for Video Technology}, vol.~35, no.~5, pp. 4870--4882, 2025.

\bibitem{cui2026depthsort}
Z.~Cui, T.~Xu, Z.~Tang, X.-j. Wu, and J.~Kittler, ``{DepthSort:} {Multi-object} tracking optimization for unreliable detection with depth information,'' \emph{IEEE Signal Processing Letters}, 2026.

\bibitem{gao2022review}
S.~Gao, K.~Yang, H.~Shi, K.~Wang, and J.~Bai, ``Review on panoramic imaging and its applications in scene understanding,'' \emph{IEEE Transactions on Instrumentation and Measurement}, vol.~71, pp. 1--34, 2022.

\bibitem{shi2023panoflow}
H.~Shi \emph{et~al.}, ``{PanoFlow:} {Learning} 360{\textdegree} optical flow for surrounding temporal understanding,'' \emph{IEEE Transactions on Intelligent Transportation Systems}, vol.~24, no.~5, pp. 5570--5585, 2023.

\bibitem{9667494}
X.~Hu, Y.~An, C.~Shao, and H.~Hu, ``Distortion convolution module for semantic segmentation of panoramic images based on the image-forming principle,'' \emph{IEEE Transactions on Instrumentation and Measurement}, vol.~71, pp. 1--12, 2022.

\bibitem{10504145}
Q.~Chang, H.~Liao, X.~Meng, S.~Xu, and Y.~Cui, ``{PanoGlassNet:} {Glass} detection with panoramic rgb and intensity images,'' \emph{IEEE Transactions on Instrumentation and Measurement}, vol.~73, pp. 1--15, 2024.

\bibitem{liu2018simple_online}
K.-C. Liu, Y.-T. Shen, and L.-G. Chen, ``Simple online and realtime tracking with spherical panoramic camera,'' in \emph{Proc. ICCE}, 2018, pp. 1--6.

\bibitem{he2021know_surroundings}
Y.~He, W.~Yu, J.~Han, X.~Wei, X.~Hong, and Y.~Gong, ``Know your surroundings: Panoramic multi-object tracking by multimodality collaboration,'' in \emph{Proc. CVPRW}, 2021, pp. 2963--2974.

\bibitem{zhang2022bytetrack}
Y.~Zhang \emph{et~al.}, ``{ByteTrack:} {Multi-object} tracking by associating every detection box,'' in \emph{Proc. ECCV}, 2022, pp. 1--21.

\bibitem{cao2023observation}
J.~Cao, J.~Pang, X.~Weng, R.~Khirodkar, and K.~Kitani, ``Observation-centric {SORT}: {Rethinking} {SORT} for robust multi-object tracking,'' in \emph{Proc. CVPR}, 2023, pp. 9686--9696.

\bibitem{liu2026s3kf}
Z.~Liu \emph{et~al.}, ``{S3KF:} {Spherical} state-space kalman filtering for panoramic {3D} multi-object tracking,'' \emph{arXiv:2603.27534}, 2026.

\bibitem{wu2024depthmot}
J.~Wu and Y.~Liu, ``{DepthMOT:} {Depth} cues lead to a strong multi-object tracker,'' \emph{arXiv:2404.05518}, 2024.

\bibitem{wang2025pd}
Y.~Wang, D.~Zhang, R.~Li, Z.~Zheng, and M.~Li, ``{PD-SORT:} {Occlusion-robust} multi-object tracking using pseudo-depth cues,'' \emph{IEEE Transactions on Consumer Electronics}, vol.~71, no.~1, pp. 165--177, 2025.

\bibitem{song2026pvdepth}
C.~Song and P.~Peng, ``{PVDepth}: {Panoramic} video depth estimation via geometry-aware spatiotemporal adaptation,'' in \emph{Proc. ICML}, 2026, pp. 114\,962--114\,986.

\bibitem{video_depth_anything}
S.~Chen \emph{et~al.}, ``Video depth anything: Consistent depth estimation for super-long videos,'' in \emph{Proc. CVPR}, 2025, pp. 22\,831--22\,840.

\bibitem{cao2025panda}
Z.~Cao \emph{et~al.}, ``{PanDA:} {Towards} panoramic depth anything with unlabeled panoramas and m{\"o}bius spatial augmentation,'' in \emph{Proc. CVPR}, 2025, pp. 982--992.

\bibitem{ai2025survey_representation}
H.~Ai, Z.~Cao, and L.~Wang, ``A survey of representation learning, optimization strategies, and applications for omnidirectional vision,'' \emph{International Journal of Computer Vision}, vol. 133, no.~8, pp. 4973--5012, 2025.

\bibitem{lin2025one_flight}
X.~Lin \emph{et~al.}, ``One flight over the gap: A survey from perspective to panoramic vision,'' \emph{arXiv:2509.04444}, 2025.

\bibitem{Zhong_2025_ICCV}
D.~Zhong \emph{et~al.}, ``{OmniSAM:} {Omnidirectional} segment anything model for {UDA} in panoramic semantic segmentation,'' in \emph{Proc. ICCV}, 2025, pp. 23\,892--23\,901.

\bibitem{chang2026dapss}
Y.~Chang, Z.~Cao, X.~Zheng, X.~Mi, and Z.~Dong, ``Denoise and align: {Towards} source-free {UDA} for robust panoramic semantic segmentation,'' in \emph{Proc. CVPR}, 2026.

\bibitem{jaus2023panoramic_panoptic}
A.~Jaus, K.~Yang, and R.~Stiefelhagen, ``Panoramic panoptic segmentation: Insights into surrounding parsing for mobile agents via unsupervised contrastive learning,'' \emph{IEEE Transactions on Intelligent Transportation Systems}, vol.~24, no.~4, pp. 4438--4453, 2023.

\bibitem{mei2022waymo}
J.~Mei \emph{et~al.}, ``Waymo open dataset: Panoramic video panoptic segmentation,'' in \emph{Proc. ECCV}, 2022, pp. 53--72.

\bibitem{fu2025panopticnerf_360}
X.~Fu \emph{et~al.}, ``{PanopticNeRF-360:} {Panoramic} {3D-to-2D} label transfer in urban scenes,'' \emph{IEEE Transactions on Pattern Analysis and Machine Intelligence}, vol.~47, no.~10, pp. 8804--8822, 2025.

\bibitem{jiang2025da360}
H.~Jiang, Z.~Song, Z.~Lou, R.~Xu, and M.~Tan, ``Depth anything in 360{\textdegree}: Towards scale invariance in the wild,'' \emph{arXiv:2512.22819}, 2025.

\bibitem{di2025hybridtrack}
L.~Di~Bella, Y.~Lyu, B.~Cornelis, and A.~Munteanu, ``{HybridTrack:} {A} hybrid approach for robust multi-object tracking,'' \emph{IEEE Robotics and Automation Letters}, vol.~10, no.~7, pp. 7238--7245, 2025.

\bibitem{yang2020using_panoramic_videos}
F.~Yang, F.~Li, Y.~Wu, S.~Sakti, and S.~Nakamura, ``Using panoramic videos for multi-person localization and tracking in a {3D} panoramic coordinate,'' in \emph{Proc. ICASSP}, 2020, pp. 1863--1867.

\bibitem{fischer2023cc_3dt}
T.~Fischer, Y.-H. Yang, S.~Kumar, M.~Sun, and F.~Yu, ``{CC-3DT:} {Panoramic} {3D} object tracking via cross-camera fusion,'' in \emph{Proc. CoRL}, 2023, pp. 2294--2305.

\bibitem{yang2026robust}
Z.~Yang \emph{et~al.}, ``Robust panoramic multi-object tracking with category-aware data association and adaptive noise estimation for unmanned surface vehicles,'' \emph{Expert Systems with Applications}, p. 132283, 2026.

\bibitem{kuhn1955hungarian}
H.~W. Kuhn, ``The hungarian method for the assignment problem,'' \emph{Naval Research Logistics Quarterly}, vol.~2, no. 1-2, pp. 83--97, 1955.

\bibitem{shen2024multi_object_tracking}
J.~Shen and H.~Yang, ``Multi-object tracking model based on detection tracking paradigm in panoramic scenes,'' \emph{Applied Sciences}, vol.~14, no.~10, p. 4146, 2024.

\bibitem{lo2022depth_aware_spherical}
L.~Lo~Presti, G.~Mazzola, G.~Averna, E.~Ardizzone, and M.~La~Cascia, ``Depth-aware multi-object tracking in spherical videos,'' in \emph{Proc. ICIAP}, 2022, pp. 362--374.

\bibitem{zhao2025detrack}
W.~Zhao, Y.~Jiang, Y.~Gao, J.~Li, and X.~Gao, ``{DETrack:} {Depth} information is predictable for tracking,'' \emph{Neurocomputing}, vol. 616, p. 128906, 2025.

\bibitem{khanchi2025depth_scoring}
M.~Khanchi, M.~Amer, and C.~Poullis, ``Depth-aware scoring and hierarchical alignment for multiple object tracking,'' in \emph{Proc. ICIP}, 2025, pp. 2043--2048.

\bibitem{yang2025depth_crowded}
C.-Y. Yang \emph{et~al.}, ``A depth-aware robust multi-object tracker for crowded scene by re-prioritizing association order,'' in \emph{Proc. AVSS}, 2025, pp. 1--6.

\bibitem{peng2025multi_densely_occluded}
J.~Peng, Y.~Yao, P.~Wang, C.~Wang, and Z.~Li, ``Multi-object tracking optimization in densely occluded scenarios using depth estimation,'' in \emph{Proc. FCN}, 2025, pp. 1--6.

\bibitem{limanta2024camot}
F.~Limanta, K.~Uto, and K.~Shinoda, ``{CAMOT:} {Camera} angle-aware multi-object tracking,'' in \emph{Proc. WACV}, 2024, pp. 6465--6474.

\bibitem{quach2024depth}
K.~G. Quach, P.~Nguyen, C.~N. Duong, T.~D. Bui, and K.~Luu, ``Depth perspective-aware multiple object tracking,'' in \emph{Engineering Applications of AI and Swarm Intelligence}, 2024, pp. 181--205.

\bibitem{sun2025view}
H.~Sun, Y.~Li, G.~Yang, Z.~Su, and K.~Luo, ``View adaptive multi-object tracking method based on depth relationship cues,'' \emph{Complex \& Intelligent Systems}, vol.~11, no.~2, p. 145, 2025.

\bibitem{han2025grasptrack}
X.~Han \emph{et~al.}, ``{GRASPTrack:} {Geometry-reasoned} association via segmentation and projection for multi-object tracking,'' \emph{arXiv:2508.08117}, 2025.

\bibitem{tran2025depthtrack}
T.~H.-P. Tran \emph{et~al.}, ``{DepthTrack:} {Cluster} meets {BEV} for multi-camera multi-target {3D} tracking,'' in \emph{Proc. ICCVW}, 2025, pp. 5348--5357.

\bibitem{piccinelli2025unik3d}
L.~Piccinelli \emph{et~al.}, ``{UniK3D:} {Universal} camera monocular {3D} estimation,'' in \emph{Proc. CVPR}, 2025, pp. 1028--1039.

\bibitem{depth_anything_v2}
L.~Yang \emph{et~al.}, ``Depth anything {V2},'' in \emph{Proc. NeurIPS}, 2024, pp. 21\,875--21\,911.

\bibitem{depthanything3}
H.~Lin \emph{et~al.}, ``Depth anything 3: Recovering the visual space from any views,'' \emph{arXiv:2511.10647}, 2025.

\bibitem{hu2025-DepthCrafter}
W.~Hu \emph{et~al.}, ``{DepthCrafter:} {Generating} consistent long depth sequences for open-world videos,'' in \emph{Proc. CVPR}, 2025, pp. 2005--2015.

\bibitem{li2025da2depthdirection}
H.~Li \emph{et~al.}, ``{DA\({}^{\mbox{2}}\):} {Depth} anything in any direction,'' \emph{arXiv:2509.26618}, 2025.

\bibitem{luiten2021hota}
J.~Luiten \emph{et~al.}, ``{HOTA:} {A} higher order metric for evaluating multi-object tracking,'' \emph{International Journal of Computer Vision}, vol. 129, no.~2, pp. 548--578, 2021.

\bibitem{bernardin2008clearmot}
K.~Bernardin and R.~Stiefelhagen, ``Evaluating multiple object tracking performance: {The} {CLEAR} {MOT} metrics,'' \emph{EURASIP Journal on Image and Video Processing}, vol. 2008, no.~1, p. 246309, 2008.

\bibitem{ristani2016idmetrics}
E.~Ristani, F.~Solera, R.~Zou, R.~Cucchiara, and C.~Tomasi, ``Performance measures and a data set for multi-target, multi-camera tracking,'' in \emph{Proc. ECCVW}, 2016, pp. 17--35.

\bibitem{schuhmacher2008consistent}
D.~Schuhmacher, B.-T. Vo, and B.-N. Vo, ``A consistent metric for performance evaluation of multi-object filters,'' \emph{IEEE Transactions on Signal Processing}, vol.~56, no.~8, pp. 3447--3457, 2008.

\bibitem{ravi2024sam2}
N.~Ravi \emph{et~al.}, ``{SAM} 2: {Segment} anything in images and videos,'' in \emph{Proc. ICLR}, 2025.

\end{thebibliography}
